\documentclass{elbioimp2}
\usepackage[utf8]{inputenc}
\usepackage{csquotes}
\usepackage[style=numeric-comp,sorting=none]{biblatex}
\usepackage{amsmath,amsfonts}
\usepackage{algorithmic}
\usepackage{algorithm}
\usepackage{array}
\usepackage[caption=false,font=normalsize,labelfont=sf,textfont=sf]{subfig}
\usepackage{textcomp}
\usepackage{stfloats}
\usepackage{url}
\usepackage{verbatim}
\usepackage{graphicx}
\usepackage{bm}

\usepackage[multiple]{footmisc}

\newcommand{\reals}{\mathbb{R}}
\newcommand{\expectation}{\mathrm{E}}
\newcommand{\expectationp}[2][]{%
    \ifx #1 \undefined \expectation{}\mathopen{}\left[#2\right]\mathclose{} % 
    \else \expectation_{#1}\mathopen{}\left[#2\right]\mathclose{} \fi}
\newcommand{\variance}{\mathrm{Var}}
\newcommand{\variancep}[2][]{%
    \ifx #1 \undefined \variance{}\mathopen{}\left[#2\right]\mathclose{} % 
    \else \variance_{#1}\mathopen{}\left[#2\right]\mathclose{} \fi}
\newcommand{\dataset}{\mathcal{D}}
\newcommand{\minibatch}{\mathcal{B}}
\newcommand{\kullb}[2]{D_{\text{KL}} \left( {#1} \, \lVert \, {#2} \right)}

\newcommand{\condbar}{\,|\,}  % The vertical bar in f(y | x). Can also use \mid to the same effect.
\newcommand{\normaldist}{\mathcal{N}}

\title{Multi-task and few-shot learning in virtual flow metering}
\shorttitle{Few-shot soft sensing}
\author{Kristian Løvland\footnote{Department of Engineering Cybernetics, Norwegian University of Science and Technology, Trondheim, Norway}\footnote{Solution Seeker AS, Oslo, Norway}\footnote{E-mail any correspondence to kristian@solutionseeker.no}, Bjarne Grimstad$^{1, 2}$ and Lars Struen Imsland$^{1}$}
\shortauthor{Løvland et al.}
\elbioimpreceived{19 Nov 2024}  
\elbioimppublished{28 Apr 2025}
\elbioimpfirstpage{1}
\elbioimpvolume{5}
\elbioimpyear{2025}
\begin{document}
\maketitle

\begin{abstract}
Recent literature has explored various ways to improve soft sensors by utilizing learning algorithms with transferability. A performance gain is generally attained when knowledge is transferred among strongly related soft sensor learning tasks. One setting where it is reasonable to expect strongly related tasks, is when learning soft sensors for separate process units that are of the same type. Applying methods that exploit transferability in this setting leads to what we call multi-unit soft sensing.

This paper formulates a probabilistic, hierarchical model for multi-unit soft sensing. The model is implemented using a deep neural network. The proposed learning method is studied empirically on a large-scale industrial case by developing virtual flow meters (a type of soft sensor) for 80 petroleum wells. We investigate how the model generalizes with the number of wells/units. We demonstrate that multi-unit models learned from data from many wells permit few-shot learning of virtual flow meters for new wells. Surprisingly, regarding the difficulty of the tasks, few-shot learning on 1-3 data points often leads to high performance on new wells.
\end{abstract}

\section{Introduction}

Many processes have variables of interest that are hard to measure. One example is a key performance indicator that cannot be measured directly, or whose measurement requires expensive equipment. Another example is a variable of interest that can only be measured by conducting a laborious or disruptive experiment, which inhibits frequent measurements. The purpose of developing \emph{soft sensors} for such processes is to make timely inferences about the variables of interest based on other process variables, whose measurements are cheaper and more accessible. 

\subsection{Soft sensing}
The literature on soft sensing is extensive and presents many applications, including online prediction, process monitoring and optimization, fault detection, and sensor reconstruction; cf. \cite{Fortuna2007,Kadlec2009,Jiang2021}. The last decade of this literature may be characterized by a stream of works that have adopted recent advancements in data science, particularly within deep learning, to develop \emph{data-driven soft sensors} \cite{Yuan2020,Sun2021,Guo2023}. Many of these works target the process industry, which has many complex processes with hard-to-measure variables and which demands the development of new and cost-efficient soft sensor technologies \cite{Ge2017,Jiang2021,Esche2022}. 

In essence, a soft sensor is a mathematical model that operates on available process measurements ($x$) to infer a variable of interest ($y$). Arguably, the most common form of a soft sensor is $\expectationp{y \condbar x, \theta} = F(x; \theta)$, where the conditional expected value of $y$ is modeled by a function $F$ with arguments $x$ and parameters $\theta$. Two important assumptions in the development of a soft sensor are i) that the process measurements, $x$, are informative of $y$, and ii) that the process measurements arrive with a higher frequency than measurements of $y$. Under these assumptions, a soft sensor can be used to monitor $y$ indirectly at times when only $x$ is measured; when $x$ is available online, a soft sensor can provide online predictions of $y$. Furthermore, when $y$ is a key performance or quality indicator, the process can be optimized based on the inferred values. 

% When applied for online prediction, a soft sensor is used to indirectly monitor $y$ at times when only $x$ is measured. Furthermore, when $y$ is a key performance indicator, the process can be optimized based on the inferred values. 

A soft sensor can be classified as being either \emph{model-driven} or \emph{data-driven}, based on the specification of $F$ \cite{Kadlec2009}. In a model-driven soft sensor, the model $F$ is derived from first-principles.
% \footnote{Arguably, most sensors are technically model-driven, since they rely on a mapping to the variable of interest from a physical quantity which is practical to process and record (e.g. an electric voltage). In the context of soft sensing, however, the term is typically limited to models which are not integrated in the sensor software.}
For example, $F$ may be a mechanistic model whose parameters represent physical properties of the process being modeled. Note that while most sensors rely on some kind of model to map the variable of interest to a physical quantity which is practical to process and record (e.g. an electric voltage), the term ``model" here refers models which are not integrated and shipped as part of the sensor.

In a data-driven soft sensor, $F$ is a generic function learned from data, e.g. a linear regression model or a neural network. Soft sensors that combine the model-driven and data-driven approaches are sometimes referred to as \emph{hybrid soft sensors}. Regardless of the soft sensor type, the model should be calibrated to available measurement pairs $\dataset = \{(x, y)\}$ before being used for prediction. Commonly used methods include least-squares or maximum likelihood estimation (MLE). When a \emph{prior} is available, maximum a posteriori (MAP) estimation or Bayesian inference can be used to calibrate the model \cite{Bishop2006}. 

Two advantages of model-driven soft sensors are: first, that they can be applied to any process to which $F$ is a valid process model; second, that they are data-efficient since the number of parameters is small. For first-principles models, the parameters are interpretable, which can allow meaningful priors to be imposed, further improving the data efficiency. One disadvantage of model-driven soft sensors is that they can be expensive or difficult to develop and maintain, especially when the modeled variable, $y$, is explained by complex phenomena. Furthermore, any model of a real process will be an approximation, and if the model is too simple, it may not have the capacity to fit the observed data, leading to local calibration and poor generalization.

The disadvantages of model-driven soft sensors have motivated much of the development of data-driven soft sensors. The attraction to data-driven soft sensing comes from its promise to simplify and reduce the cost of modeling by learning the model from historical data using statistical inference or machine learning techniques. A challenge with data-driven soft sensing is to learn a model from information-poor data, where the paucity of information may be due to a low data volume, frequency, variety, or quality \cite{Kadlec2009}.
The problem of low data variety in particular provides a challenge for data-driven methods, since these can generally not be expected to extrapolate outside of their training distribution. These data challenges are closely tied to the motivation for implementing a soft sensor, which is to infer a key variable that is measured infrequently. Put differently, \emph{a soft sensor is most valuable in information-poor environments, where one would expect data-driven methods to perform poorly}. To achieve a good predictive performance in such environments, it is imperative to be \emph{data efficient}. Model-driven soft sensors achieve this by having few parameters and strong priors. On the other hand, data-driven soft sensors, which may have more model parameters than observations to learn from, often rely on generic regularization techniques and weak priors. In effect, this means that their model is learned from scratch for each application. Thus, we argue that data efficiency is a pressing issue in data-driven soft sensing.

\subsection{Multi-unit soft sensing}
One strategy to making data-driven soft sensors more data efficient is to exploit \emph{transferability} \cite{Curreri2021}. A general description of transferability is the ability of a learning algorithm to transfer/share knowledge between different learning tasks to generalize better. In soft sensing, transferability entails learning across multiple soft sensor tasks. If the tasks are strongly related (i.e., share important characteristics), one may expect transferability to help soft sensors generalize. Various ways to employ transferability in soft sensing are covered in the following, where we elaborate on how different learning methods have been used to transfer knowledge between different soft sensor tasks. 

\begin{figure*}[bt]
\centering
\includegraphics[width=0.8\linewidth]{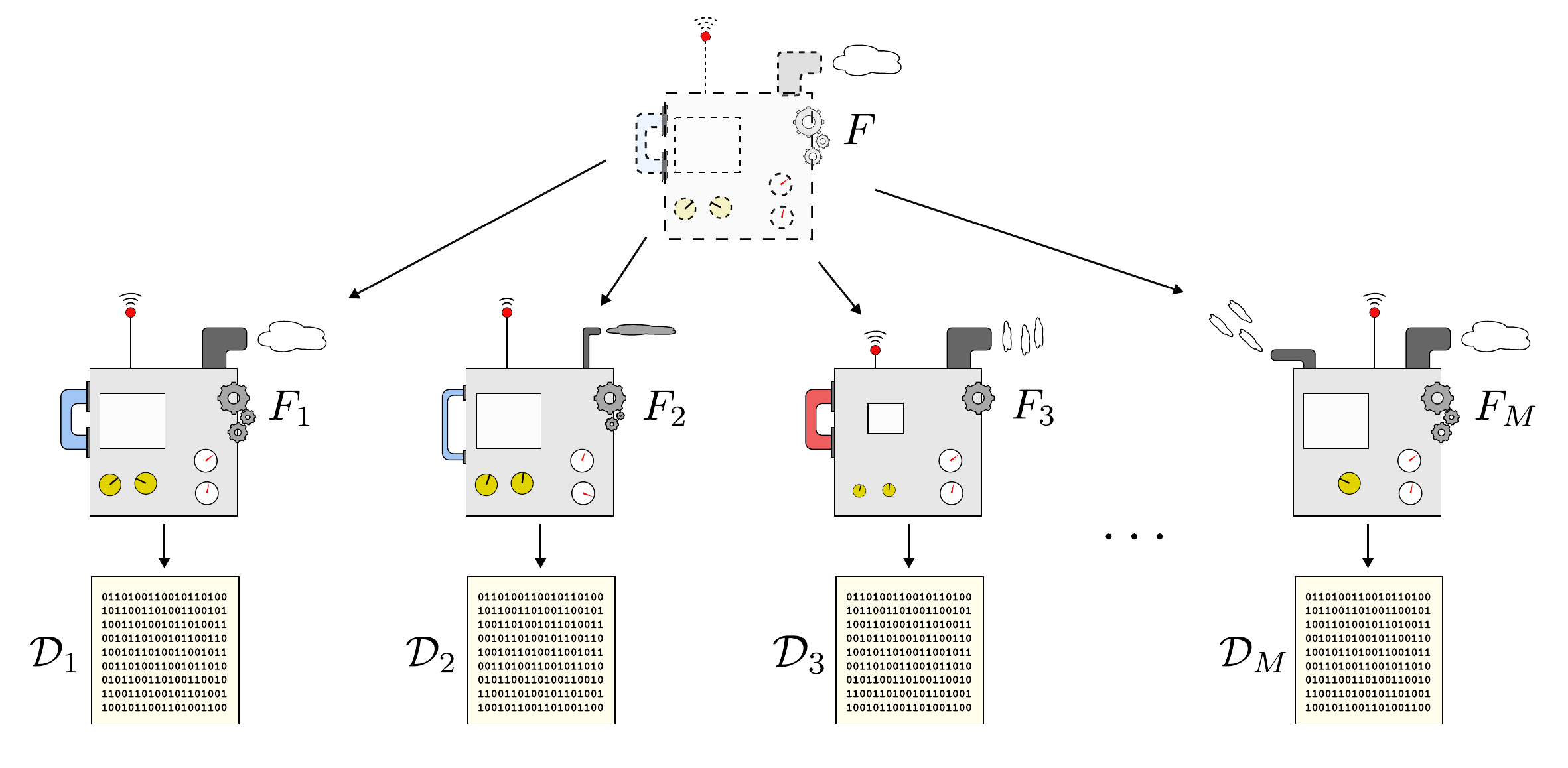}
\caption{Illustration of multi-unit data generation. Here, each unit $F_i$ corresponds to a physical unit performing the same basic operation. The $M$ units are similar to each other, and can be thought of as different realizations of some ``prototype" unit, denoted here by $F$. Still, each unit also has its own unique physical characteristics, resulting in datasets $\dataset_i$ with different distributions.}
\label{fig:multi-unit}
\end{figure*}

A particularly relevant case for transferability is when the same type of soft sensor is to be developed for several similar, but physically different processes. An example is when a product quality indicator (soft sensor) is to be developed for several, almost identical production lines. In this case, it is reasonable to expect the different soft sensor tasks to be strongly related. We call the application of transferability in this setting \emph{multi-unit soft sensing}, where \emph{unit} refers to a physical process being modeled by a soft sensor.\footnote{In other works, a unit may be referred to as an entity, subject, object, or realization for which we have multiple observations. In statistical terms, it would be called a \emph{unit of analysis} with repeated measurements. Also note that a unit may represent a \emph{unit operation}, which is a term used in chemical engineering to refer to a basic step in a larger process.} For example, \emph{units} may refer to a set of pumps, valves, solar panels, distillation columns, geothermal wells, or robots. The setting of multi-unit soft sensing is illustrated in Figure \ref{fig:multi-unit}.

Data-driven multi-unit soft sensing will be framed in mathematical terms in our problem statement. Interestingly, the framing allows us to draw a parallel to model-driven soft sensing. Suppose that we have access to a large multi-unit dataset, in terms of the number of observations and units. One could then expect that a learning algorithm with transferability will converge to a prototypical representation of the data (which we can view as a strong prior model). When facing new units, this prototypical representation can be leveraged to quickly learn a soft sensor from a few data points. In machine learning terminology, this would be an example of \emph{few-shot learning}, where a model performs well on a task after being calibrated to a few (often less than ten) data points \cite{Finn2017,Zintgraf2019}. The parallel to model-driven soft sensing is that a first-principles model and a learned prototypical model take the same role of representing the general process. 

\subsection{Virtual flow metering}
Virtual flow metering is a widely studied soft sensing problem, where fluid flow rates are inferred from process measurements \cite{Bikmukhametov2020}. In petroleum production, virtual flow metering is extensively used to gauge the flow rates of oil, gas and water in wells. The multi-phase flow rates inform many important operating decisions and safety assessments. Because of this, virtual flow metering is an integral part of the operating practices of many petroleum assets.

Virtual flow metering technologies are valuable because measuring multi-phase flow rates is a complex problem. In petroleum production, flow rates are typically measured using a \textit{test separator}, shared by many wells. Using a test separator involves an intervention: routing a well to the separator, waiting for transient behaviors to settle, and then performing the measurement. This is a potentially disruptive operation that spans several hours. While it may result in accurate measurements, it often leads to measurements taken under different-from-normal operating conditions. With this setup, the flow rates of a well are typically measured every 7-60 days. Flow rates can also be measured by a \textit{multi-phase flow meter} (MPFM), which can be installed separately on each well. MPFMs provide continuous and non-invasive, but typically less accurate flow rate measurements. These are complex devices that tend to suffer from sensor drift and require regular calibration to work well \cite{Hansen2019}. MPFMs are expensive and therefore not installed on all wells. 

Virtual flow meters (VFMs) based on first principles models (e.g. \cite{schlumberger2014dynamic, technipfmc2024flowmanager, petex2024prosper, kongsberg2024ledaflow}) are currently favored by the industry. These VFMs have demonstrated that they can be accurate when properly calibrated, even in data-scarce settings. Still, first principles VFMs are subject to the typical limitations of model-driven soft sensors. In particular, the high cost of setting up and maintaining such VFMs risks limiting their performance (due to insufficient maintenance). The high cost also prohibits their use on assets with small investment budgets.

Data-driven VFMs attempt to address the cost issue with first principles VFMs by lowering the model setup and maintenance costs. In addition to reducing manual labor related to physical modeling, data-driven VFMs typically also reduce the need for manual parameter tuning by allowing for automatic model tuning in the presence of new data \cite{Nemoto2023, Akhiiartdinov2023}.

If the flow conditions are sufficiently predictable, data-driven models based on simple first principles equations (which need to be simple enough to allow for automatic estimation of all model parameters from data) may yield satisfactory prediction accuracy \cite{Staff2020}. If prior knowledge about flow conditions is lacking, it may be possible to compensate by learning from more data; in such settings, steady-state VFMs based on deep learning have shown promising results \cite{Al2018, Sandnes2021, Hotvedt2022, Bikmukhametov2020}. In the presence of data with high time resolution, methods based on deep neural networks have also been shown capable to model dynamical flow phenomena like slugging and transient behavior following step changes \cite{Andrianov2018}.

It is commonly believed that the successful use of data-driven VFMs based on deep neural networks requires large amounts of data \cite{Bikmukhametov2020}. Such a requirement would limit their applicability, since the wells with the worst instrumentation (and hence, the least data) are typically also the wells were VFMs are needed the most. Thus, the application of data-driven VFM to wells with few observations is a problem of high interest.

\subsection{Contributions}
We provide a probabilistic, hierarchical model formulation of multi-unit soft sensing, where parameters enter the model at different levels to capture variations among units and observations. Nonlinearities are represented by a deep neural network with an architecture that is inspired by recent works in meta learning \cite{Zintgraf2019} and multi-task learning \cite{Sandnes2021}. The architecture features learned, unit-specific parameters that modulate the represented nonlinearities to enable unit adaptation.

We investigate the learning capabilities of the multi-unit soft sensing model on a large-scale, industrial VFM case. The main findings are listed below.
\begin{itemize}
    \item The average soft sensor performance (across units) increases with the number of units. The convergence rate matches a theoretical estimate, and can be used to indicate when a good base model is found.
    \item A base model trained on many units, enables few-shot learning on other units. In our case, calibration of unit-specific parameters to only one or two observations leads to a low prediction error for new units. 
\end{itemize}
To our knowledge, we are the first to investigate the few-shot learning capabilities of a multi-unit soft sensing model. 

Our findings are relevant for industrial soft sensor applications, and in particular data-driven VFMs. First, the scaling properties indicate that soft sensor generalizability may be improved by using multi-unit data. Second, the demonstrated few-shot learning capability enables the application of data-driven soft sensors in data-scarce settings. For units with few observations, many have considered model-driven or hybrid soft sensors as the only suitable solutions; see e.g. \cite{Bikmukhametov2020,Hsiao2021}. Our finding challenges this view. It also suggests that purely data-driven soft sensors can be run and calibrated (or fine-tuned) on edge computing hardware, where computational resources are limited \cite{Shi2016}. This may enable manufacturing or service companies that want to bundle data-driven soft sensors with their hardware offers.

\subsection{Paper outline}
The paper proceeds by discussing related works , before giving a mathematical problem formulation. The multi-unit soft sensor model is then described, followed by the derivation of the learning method. An empirical study based on a large-scale industrial dataset is presented, before concluding remarks are given. Some implementation details and additional results are gathered in an appendix.

\subsection{Notation} 
A parametric function $f$ with inputs $(x, y)$ and parameters $(a, b)$ is denoted by $f(x, y; a, b)$, where the input and parameter arguments are separated by a semicolon. 

Sets of variables are compactly represented by bold symbols, e.g. $\bm{x} = \{x_i\}_{i=1}^{N}$, where $x_i \in \reals^D$ for some positive integers $D$ and $N$. For double-indexed variables we write $\bm{x} = \{x_{ij}\}_{ij}$ when the index sets of $i$ and $j$ are implicit. When the indices are irrelevant to the discussion, we may write $x$ instead of $x_{ij}$.

For some positive integer $K$, we denote the $K$-vector of zeros by $0_K$ and the $K\times K$ identity matrix by $I_K$. For a $K$-vector $\sigma = (\sigma_1, \ldots, \sigma_K)$, $\log(\sigma)$ denotes the element-wise natural logarithm, $\exp(\sigma)$ the element-wise natural exponential, and $\text{diag}(\sigma^{2})$ denotes the diagonal $(K\times K)$-matrix with diagonal elements $(\sigma_{1}^{2}, \ldots, \sigma_{K}^{2})$. 

The normal distribution is denoted by $\normaldist(\mu, \Sigma)$, where $\mu$ is a mean vector and $\Sigma$ is a covariance matrix. If $X$ is a normally distributed random variable, we denote a sample by $x \sim \normaldist(x \condbar \mu, \Sigma)$. Generally, we use the letter $p$ to denote a probability distribution, $p(x)$, or a conditional probability distribution, $p(y \condbar x)$. We sometimes use the letter $q$ for probability distributions that are approximative.

\section{Related works}
\label{sec:related-works}
A recent theme in the soft sensing literature is transferability, or transfer learning \cite{Curreri2021,Sun2021,Zhai2023}. Simply put, transfer learning entails taking knowledge acquired by solving a source task and utilizing it to improve generalization on a target task \cite{Pan2010}. Here, a task would be to develop a soft sensor. The knowledge, captured by a model, is then transferred from one soft sensor (source) to the next (target). The model learned on the source task may act as a stronger prior than a generic or uninformative prior on the target task, and in this way improve data efficiency.

The \emph{multi-task learning} (MTL) paradigm generalizes the unidirectional transfer (described above) to multidirectional transfer between two or more tasks \cite{Zhang2021}. In the MTL paradigm, multiple soft sensors are learned simultaneously and knowledge transfer happens through hard or soft parameter-sharing \cite{Zhai2023,Huang2023}. A common way to implement MTL with hard parameter-sharing in deep neural networks, is to share all of the hidden layers, and to have one output layer per task \cite{Zhang2021}. This results in a high number of shared parameters, and task-specific parameters that are lower-dimensional. This setup may be beneficial when there are many similar tasks, but few observations per task.

Transferability, whether it happens via transfer learning or MTL, can be used in different ways to improve soft sensors. Here, we discuss recent attempts to employ transferability along four ``dimensions'': across data fidelity (e.g. from synthetic to real data), across domains (e.g. operating conditions), across target variables (e.g. quality indicators), and across processes (e.g. similar but physically different processes).

In \cite{Hsiao2021}, a soft sensor was developed by first training on synthetic data generated by a first-principles simulator, and then fine-tuning the model on a limited amount of real process data. In this case, transfer learning provided a mechanism for combining real and synthetic data to improve the model. A draw-back with the approach is that it requires a first-principles simulator.

It is often the case that the operating practices, operating conditions, or dynamics of industrial processes induce a data shift, e.g. a covariate shift, meaning that the source-domain data is differently distributed than the target-domain data. Transfer learning across different operating conditions is referred to as \emph{cross-phase transfer learning} in \cite{Curreri2021}. As demonstrated in \cite{Chai2022,zhang_online_2023,zhang_gaussian_2023}, cross-phase transfer learning can enhance the soft sensor performance on the target domain. Operating conditions can also change due to sensor faults. Fault-tolerant soft sensors can then be developed using techniques from domain adaptation, a subcategory of transfer learning \cite{Zhang2023}.

In many processes there are multiple target variables, e.g. different key performance indicators or quality indicators, for which soft sensors are developed. When these target variables are related, it may be beneficial to transfer learn across the soft sensors \cite{Liu2019,Yan2020,Qiao2023,Huang2023}. As an example, consider a manufacturing plant which does not work as intended, leading to an overall poor product quality and correlation among quality indicators. 

Finally, knowledge can be transferred between soft sensors for similar, but physically different processes. This setting is referred to as \emph{cross-entity transfer learning} in \cite{Curreri2021}. The idea behind cross-entity transfer learning, is that, if two entities/processes/units are physically similar, data from one unit will be informative to the operation of the other unit. Thus, with data from multiple units, soft sensor performance may be improved by pooling the data and learning across units. This learning paradigm has high potential since many soft sensors are developed for processes that are commonly found in the industry; consider a soft sensor for a specific pump unit, which may be of a model produced in large numbers and installed in many process systems. Despite the high potential, few examples of cross-entity transfer learning can be found in the literature (one reason may be that multi-unit datasets are rarely shared). One example is given in \cite{Sandnes2021}, where a data-driven virtual flow meter is learned from data from many oil wells using multi-task learning.

\subsection{Nomenclature}
In our work, we prefer to use the term \emph{multi-unit soft sensing} since it reflects that learning happens across units without specifying by which means (e.g. transfer learning or multi-task learning). In settings where the means of transferability is relevant, we will however state which paradigm best describes the learning problem in question. Briefly summarized, \textit{pretraining} will be done through the use of multi-task learning while \textit{few-shot learning} will be done through a finetuning/calibration strategy which constitutes a form of transfer learning. We will use the term \textit{few-shot transfer learning} in settings where we believe this deserves emphasis. In the experimental case study we will apply cross-entity transfer learning; the terms \textit{unit} and \textit{task} will be used interchangeably in this particular context.

\section{Problem statement}
\label{sec:problem-statement}
Consider a set of $M > 1$ distinct, but related \emph{units} indexed by $i \in \{1, \ldots, M\}$. The units are assumed to be related so that it is reasonable to expect, a priori to seeing data, that transfer learning between units is beneficial. For each unit $i$, we have at our disposal a set of $N_i$ observations $\dataset_{i} = \{(x_{ij}, y_{ij})\}_{j=1}^{N_i}$, where $x_{ij} \in \reals^{D}$ denotes explanatory variables and $y_{ij} \in \reals$ are target variables. We assume that the variables represent similar measurements or quantities across units. We also assume that each dataset $\dataset_i$ consists of independent and identically distributed (iid.) observations drawn from a probability distribution $p_i(x,y) = p_i(y \condbar x) p_i(x)$ over $\reals^{D} \times \reals$. 

Our goal is to efficiently learn the behavior of the units, as captured by the conditional densities $p_1(y \condbar x), \ldots, p_M(y \condbar x)$, from the data collection $\dataset = \{\dataset_1, \ldots, \dataset_M\}$. In particular, we wish to exploit that some structures or patterns are common among units. 

Invariably, distinct units will differ in various ways, e.g., they may be of different design or construction, they may be in different condition, or they may operate under different conditions. The properties or conditions that make units distinct are referred to as the \emph{context} of a unit. An observed context may be included in the dataset as an explanatory variable. An unobserved context must be treated as a latent variable to be inferred from data and it is only possible to make such inferences by studying data from multiple, related units. We assign the letter $c$ to latent context variables.

We consider the underlying process $p(x, y)$ that generates data for any unit in a population. Let $c \in \reals^K$ be a latent context variable with distribution $p(c)$. The context identifies a unit in the population supported by $p(c)$. We can then think of the process as the marginalization over all units, i.e., $p(x, y) = \int p(x, y \condbar c) p(c) dc$. With this model, $p(x, y \condbar c_i) = p_i(x, y)$, for a given realization $c = c_i$. This implies that $p(y \condbar x, c_i) = p_i(y \condbar x)$, showing that we can model the behavior of all units by a single conditional distribution. 

We now state that the goal of multi-unit soft sensing is to learn from $\dataset$ a distribution $q(y \condbar x, c)$ that approximates $p(y \condbar x, c)$. We contrast this to the single-unit soft sensing, where $p_i(y \condbar x)$ is approximated using $\dataset_i$, for $i=1,\ldots,M$; that is, where $M$ soft sensors are learned separately. 

Our work aims to study empirically and theoretically the predictive performance of the approximation $q(y \condbar x, c)$. First, we want to investigate how performance scales with the number of units $M$. Second, we want to examine the few-shot learning performance of $q(y \condbar x, c)$ as the context $c$ is calibrated to few observations of a new unit (whose data was not used to learn $q$).

\section{Model description}
\label{sec:model}
\begin{figure}[ht]
\centering
\includegraphics[width=0.6\linewidth]{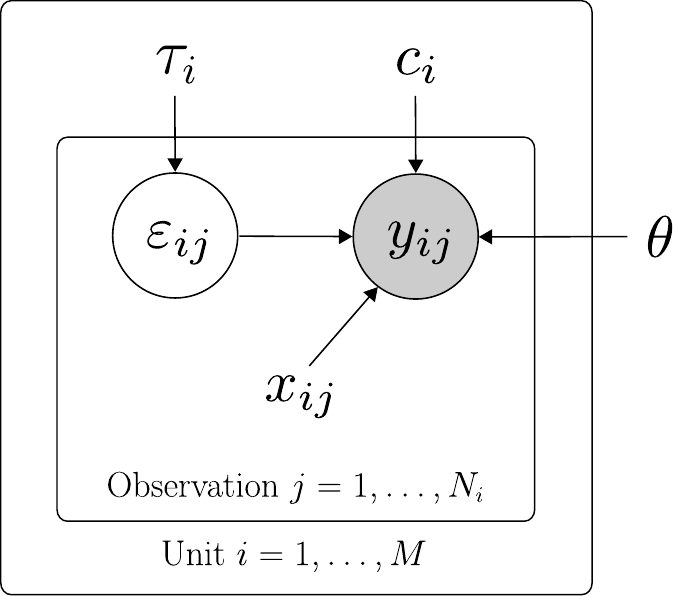}
\caption{Multi-unit soft sensing model. Random variables are encircled. A grey (white) circle indicates that the variable is observed (latent). The nested plates (rectangles) group variables at different levels.}
\label{fig:mtl-model}
\end{figure}

We consider a hierarchical model of the observed target variables for a set of units. The model has three levels, as illustrated graphically in Figure \ref{fig:mtl-model}. The top-level has \emph{universal} parameters ($\theta$) that are shared by all units and observations. The middle \emph{unit layer} has parameters ($c$ and $\tau$) that capture the specificity of units. At the inner \emph{observation level}, the target variable ($y$) is explained by the explanatory variables ($x$) and the context ($c$) of the observed unit. The target observation is subject to a measurement error ($\varepsilon$) with a unit-specific variance ($1/\tau$). The relationship between these variables is modeled using the universal parameters ($\theta$). The hierarchical structure allows the model to capture variations at the level of units and observations. A mathematical description of the model is given next.

For unit $i \in \{1, \ldots, M\}$, we model observation $j \in \{1, \ldots, N_i\}$ as
\begin{equation*}
    y_{ij} = f(x_{ij}; c_i, \theta) + \varepsilon_{ij},
\end{equation*}
where $\varepsilon_{ij}$ is the error term. The error, here interpreted as measurement noise, is modeled as $\varepsilon_{ij} \sim \normaldist(0, 1 / \tau_i)$, where the precision (or variance $\sigma_i^2 = 1/\tau_i$) is allowed to differ between units. The function $f$ is modeled by a neural network with parameters $(c, \theta)$ and arguments $x$. The specific architecture of $f$ is described in the appendix.

We can express the observation model as a conditional probability density:
\begin{equation}
\begin{aligned}
    y_{ij} \condbar x_{ij}, c_{i}, \tau_i, \theta &\sim p(y_{ij} \condbar x_{ij}, c_{i}, \tau_i, \theta) \\
    &= \normaldist(y_{ij} \condbar f(x_{ij}; c_i, \theta), 1/ \tau_i).
    \label{eq:obs-model}
\end{aligned}
\end{equation}

The model in \eqref{eq:obs-model} results from several simplifying assumptions: 1) the dimension of the context variables, $K$, is fixed and considered a design choice; 2) the model is conditional on $x$, and thus not a generative model of the observed data; 3) the noise is Gaussian; 4) the model is homoscedastic since the measurement noise variance, $\sigma_i^2$, is fixed (per unit). Assumption 2 leads to a discriminative model which is suitable when the soft sensor will be used for on-line prediction. Assumption 3 could be relaxed (e.g. by generalization to exponential family) without much technical complication, but is kept for reasons of simplicity. Relaxing assumption 4, for example by allowing $\sigma_i$ to vary with $x$, would give a more complex heteroscedastic model, which may be harder to learn in a data-scarce setting.

\subsection{Priors}
\label{sec:model-priors}
The neural network parameters $\theta$ are universal, i.e., shared by all units. We want the function $f$ to behave smoothly, so that the effect of varying the unit-specific context parameters, $c$, leads to a predictable pattern. To regularize the neural network, we may impose on $\theta$ a Gaussian prior, $p(\theta) = \normaldist(0, \Sigma_\theta)$. This prior corresponds to an $L^2$-regularization of the $\theta$-parameters \cite{Goodfellow2016}.

The context parameters, $\bm c$, are used to modulate $f$ to fit the data of each unit. By putting a common prior on these parameters, we regularize the parameter-space. In this work, we use a common standard normal prior $p(c_1) = \ldots = p(c_M) = \normaldist(0_K, I_K)$. The identity covariance matrix is justified since the neural network $f$ has the capacity to scale the context variables. Pinning the covariance matrix to the identity matrix simplifies modeling.

On the precision parameters, $\tau_i$, we put a gamma prior $p(\tau_i) = \text{Gamma}(\alpha, \beta)$, where the hyper-parameters $\alpha$ and $\beta$ are the concentration and rate, respectively. This prior corresponds to a zero-avoiding inverse-gamma prior on the variance \cite{Gelman2014}. For a normalized target variable $y$, we specify a weakly informative prior on the precision by setting $\alpha = 1$ and $\beta = 0.001$. This setting gives a prior on the precision with expected value $10^3$ and variance $10^6$.

The prior on $(\bm c, \bm \tau, \theta)$ can be factorized as follows:
\begin{equation}
    \log p(\bm c, \bm \tau, \theta) = \sum\limits_{i=1}^{M} \left[ \log p(c_i) + \log p(\tau_i) \right] + \log p(\theta).
    \label{eq:log-priors}
\end{equation}

\subsection{Likelihood}
\label{sec:model-likelihood}
We collect the variables in the following sets to allow a compact notation: $\bm{x} = \{x_{ij}\}_{ij}$, $\bm{y} = \{y_{ij}\}_{ij}$, $\bm{c} = \{c_{i}\}_{i}$, and $\bm{\tau} = \{\tau_{i}\}_{i}$. The log-likelihood can then be expressed as:
\begin{equation}
\begin{aligned}
    \ell(\bm c, \bm \tau, \theta \condbar \dataset) :=& \log p(\bm y \condbar \bm x, \bm c, \bm \tau, \theta) \\ =& \sum\limits_{i=1}^{M} \sum\limits_{j=1}^{N_{i}} \log p(y_{ij} \condbar x_{ij}, c_{i}, \tau_i, \theta).
    \label{eq:log-likelihood}
\end{aligned}    
\end{equation}
We use the short-form notation $\ell(\bm c, \bm \tau, \theta \condbar \dataset)$ for the log-likelihood when it is considered as a function of the parameters ($\bm c$, $\bm \tau$, $\theta$), given data $\dataset = (\bm x, \bm y)$. We will show shortly how to perform MAP estimation of the parameters, given the log-likelihood in \eqref{eq:log-likelihood} and log-prior in \eqref{eq:log-priors}.

\section{Learning method}
\label{sec:inference-method}
In model learning, the posterior density of the parameters $(\bm c, \bm \tau, \theta)$ is of chief interest. The application of Bayes' theorem yields:
\begin{equation}
    p(\bm c, \bm \tau, \theta \condbar \bm x, \bm y) \propto p(\bm y \condbar \bm x, \bm c, \bm \tau, \theta) p(\bm c, \bm \tau, \theta),
    \label{eq:posterior}
\end{equation}
where the posterior on the left-hand side is proportional to the joint density. The joint density is the product of the prior in \eqref{eq:log-priors} and likelihood in \eqref{eq:log-likelihood}.

A maximum a posteriori (MAP) estimate of the parameters is sought by solving the following optimization problem:
\begin{equation}
    (\hat{\bm c}, \hat{\bm \tau}, \hat{\theta}) = \underset{\bm c, \bm \tau, \theta}{\arg \max} ~ p(\bm c, \bm \tau, \theta \condbar \bm x, \bm y).
    \label{eq:map-estimation}
\end{equation}
The point estimate, $(\hat{\bm c}, \hat{\bm \tau}, \hat{\theta})$, is of the mode of the parameters' posterior density. Next, we discuss how to solve this problem using a stochastic gradient ascent method. Our procedure follows standard practices in deep learning \cite{Goodfellow2016}, with some adaptations to our multi-unit soft sensor model.

\subsection{Optimization}
\label{sec:optimization}
In practice, MAP estimation is usually implemented as the maximization of an alternative objective function, here written as
\begin{equation}
    J(\bm c, \bm \tau, \theta \condbar \dataset) := \ell(\bm c, \bm \tau, \theta \condbar \dataset) + \log p(\bm c, \bm \tau, \theta).
    \label{eq:objective}
\end{equation}
This objective function is obtained by taking the logarithm of the posterior in \eqref{eq:posterior} and dropping the constant evidence term. These operations do not alter the solution (estimate), but make the objective simpler to optimize numerically. Note that all the terms making up $J$ can be computed analytically for the densities specified here.

\subsection{Constrained precision estimation}
The optimization problem above involves the implicit constraint that the target precision must be strictly positive, i.e. $\tau_i > 0$. To incorporate this constraint into an unconstrained gradient-based method, we reparameterize $\tau_i$ as follows. Let $\tau_i = g(t_i) = \log(1 + \exp(t_i))$, where $t_i \in \reals$ is a new parameter. The reparameterization uses the softplus function $g$, which maps $\reals \rightarrow \reals_{>0}$. The softplus function is injective and does not change the solution.

We denote the reparameterized objective function by
\begin{equation}
    J^r(\bm c, \bm t, \theta \condbar \dataset) := \ell(\bm c, g(\bm t), \theta \condbar \dataset) + \log p(\bm c, g(\bm t), \theta),
    \label{eq:objective-reparameterized}
\end{equation}
where $\bm t = \{t_i\}_i$ and $g(\bm t)$ is the softplus function applied element-wise to $\bm t$.

\subsection{Stochastic gradients}
Gradient-based optimization requires us to compute gradients of $J^r$ in \eqref{eq:objective-reparameterized}. Let $\nabla J^r(\bm c, \bm t, \theta \condbar \dataset)$ denote the gradient of $J^r$ with respect to the parameters $(\bm c, \bm t, \theta)$ at some point in the parameter space (here unspecified). The gradient is conveniently computed using the back-propagation algorithm \cite{Rumelhart1986}. However, the computation becomes expensive for large datasets since the likelihood has $|\dataset| = N = N_1 + \ldots + N_M$ terms. To reduce the computational burden we resort to stochastic gradient estimates.

Let $\minibatch = (\minibatch_1, \ldots, \minibatch_M)$ be a collection of randomly selected mini-batches $\minibatch_i \subset \dataset_i$ of size $B_i = |\minibatch_i|$. The log-likelihood of task $i$ can then be approximated as follows:
\begin{align*}
    \sum_{x, y \in \dataset_i} \log p(y \condbar x, c_i, \tau_i, \theta) \simeq \frac{N_i}{B_i} \sum_{x, y \in \minibatch_i} \log p(y \condbar x, c_i, \tau_i, \theta).
\end{align*}
Using this, we can obtain a mini-batch approximation of $J^r(\bm c, \bm t, \theta \condbar \dataset)$:
\begin{align*}
    J^r(\bm c, \bm t, \theta \condbar \minibatch) &= \sum\limits_{i=1}^{M} \frac{N_i}{B_i} \sum\limits_{(x, y) \in \minibatch_i} \log p(y \condbar x, c_{i}, g(t_i), \theta) \\
    &\quad + \sum\limits_{i=1}^{M} \log p(c_i) + \sum\limits_{i=1}^{M} \log p(g(t_i)) + \log p(\theta).
\end{align*}
%If $\minibatch_i = \emptyset$, we set $\log p(c_i) = \log p(g(t_i)) = 0$ in the above sum, effectively removing any contribution from unit $i$ to the objective. 
A mini-batch stochastic gradient estimate $\nabla J^r(\bm c, \bm t, \theta \condbar \minibatch)$ can then be computed by using back-propagation. Provided that the mini-batches are selected randomly (with substitution), the stochastic gradient is an unbiased estimate of $\nabla J^r(\bm c, \bm t, \theta \condbar \dataset)$. Pseudocode for the estimator is given in Algorithm \ref{alg:stochastic-gradient}. 

\begin{algorithm}[ht]
\caption{Stochastic gradient estimator}
\label{alg:stochastic-gradient}
\begin{algorithmic} % [1], for numbered lines 
\REQUIRE data $\dataset$, model $p(\bm y, \bm c, \bm \tau, \theta \condbar \bm x)$, and parameter values $(\bm c, \bm t, \theta)$. 
\STATE Randomly draw a mini-batch $\minibatch$ from $\dataset$
\STATE $\tilde{J} \gets 0$ 
\FOR{$i=1,\ldots,M$} %\Comment{Loop over units}
% \IF{$\minibatch_i \neq \emptyset$}
\FOR{$(x, y) \in \minibatch_i$} %\Comment{Loop over data}
\STATE $\tilde{J} \gets \tilde{J} + (N_i / B_i) \log p(y \condbar x, c_i, g(t_i), \theta)$
\ENDFOR
\STATE $\tilde{J} \gets \tilde{J} + \log p(c_i) + \log p(g(t_i))$ %\Comment{Computed analytically}
% \ENDIF
\ENDFOR
\STATE $\tilde{J} \gets \tilde{J} + \log p(\theta)$ %\Comment{Computed analytically}
\STATE Compute $\nabla \tilde{J}$ using back-propagation \\
\RETURN $\nabla \tilde{J}$
\end{algorithmic}
\end{algorithm}

\subsection{Mini-batch gradient ascent}
Bringing the above considerations together, we see that the original MAP estimation problem in \eqref{eq:map-estimation} can be reformulated to

%We bring the above considerations together and present a mini-batch gradient ascent method for the MAP estimation problem in \eqref{eq:map-estimation}. We optimize the reformulated problem
\begin{equation}
    (\hat{\bm c}, \hat{\bm t}, \hat{\theta}) = \underset{\bm c, \bm t, \theta}{\arg \max} ~ J^r(\bm c, \bm t, \theta \condbar \dataset),
    \label{eq:map-estimation-reformulated}
\end{equation}
where the reformulated objective is given by \eqref{eq:objective-reparameterized}. With the mini-batch gradient estimates computed by Algorithm \ref{alg:stochastic-gradient}, MAP estimation can be performed by a stochastic gradient method, such as SGD, AdaGrad or Adam \cite{Bottou2018}.

Here, we illustrate the procedure using a simple mini-batch gradient ascent approach. Starting with an initial guess for the parameters, $(\bm c, \bm t, \theta)_0$, we iteratively update the parameter values as
\begin{align}
    (\bm c, \bm t, \theta)_{k+1} = (\bm c, \bm t, \theta)_{k} + \lambda_k \nabla J^r(\bm c, \bm t, \theta \condbar \minibatch)|_{(\bm c, \bm t, \theta)_{k}}, 
    %|_{(\bm c, \bm t, \theta) = (\bm c, \bm t, \theta)_{k}},
    \label{eq:sgd}
\end{align}
where $\lambda_k$ is the learning rate at step $k$ and the gradient estimate is computed using Algorithm \ref{alg:stochastic-gradient}. Note that the mini-batch is randomly selected at each iteration. The optimization is stopped when some termination criterion is met, and the final parameter values are returned as the estimate $(\hat{\bm c}, \hat{\bm \tau}, \hat{\theta})$, where the precision estimate is retrieved as $\hat{\bm \tau} = g(\hat{\bm t})$. 

\subsection{Context calibration}
\label{sec:context-calibration}
The previous sections describe how to learn a model from data $\dataset$ collected from $M$ units.  Here, we provide a method for calibrating a learned model to data from new units; i.e., data that was not included in $\dataset$ and that we get access to after having learned the initial model. 

Let $(\hat{\bm c}, \hat{\bm \tau}, \hat{\theta})$ be the parameters estimated from $\dataset$. We assign the index $M+1$ to a new unit with data $\dataset_{M+1}$. A simple calibration procedure for $c_{M+1}$ and $\tau_{M+1}$ is to apply the gradient update in \eqref{eq:sgd} with a fixed $\theta = \hat{\theta}$ and $\minibatch = \dataset_{M+1}$. With this procedure, $K+1$ parameters are calibrated to $|\dataset_{M+1}|$ data points. Provided that we do not have any prior knowledge about unit $M+1$, it is natural to initialize $c_{M+1} = 0$ and $\tau_{M+1} = (\hat{\tau}_1 + \cdots + \hat{\tau}_M) / M$. 

The above calibration procedure works under the assumptions that $|\dataset_{M+1}| \ll |\dataset|$ and that the $\theta$ estimate has converged. When $|\dataset_{M+1}|$ contains very few data points (relative to the task to be solved), the calibration procedure can be thought of as a few-shot learning method working on a pretrained model.

In the case where $|\dataset_{M+1}|$ contains very few data points, it may also make sense to fix $\tau_{M+1}$ during training to reduce the number of learned parameters. 

\subsection{Context parameter analysis}
\label{sec:context-parameter-analysis}
The context parameters in $\bm c$ represent latent variables, and cannot be expected to exactly represent physical quantities. However, one may still hope that the learned context parameter space possess certain properties which are commonly held by physical parameter spaces. An important such property is that similar units are represented by similar context parameters. To investigate the validity of beliefs like this, methods for analyzing the context parameters are required. In this section we describe two approaches which may indicate whether the learned context parameter space is well-behaved.

\subsection{Information gain estimation}
MAP estimation, as presented above, gives a point estimate $(\hat{\bm c}, \hat{\bm \tau}, \hat{\theta})$ of the mode of the posterior distribution $p(\bm c, \bm \tau, \theta \condbar \dataset)$. Thus, it does not inform us of the uncertainty in the estimate. In the few-shot learning setting, the marginalized posterior distribution $p(\bm c \condbar \dataset)$ may provide useful information about the context uncertainty given few observations, which may, in turn, be useful for analyzing context parameters. While the exact posterior is hard to compute, a local approximation $q(\bm{c} \condbar \dataset)$ of $p(\bm{c} \condbar \dataset)$ can be obtained by Laplace's approximation \cite{Bishop2006}:
\begin{equation}
\begin{aligned}
    q(\bm c \condbar \dataset) &= \normaldist(\hat{\bm c}, S_c), \\
    S_c^{-1} & = - \nabla_{\bm c} \nabla_{\bm c} J(\bm c, \hat{\bm \tau}, \hat{\theta} \condbar \dataset) |_{\bm c = \hat{\bm c}}.
\end{aligned}
\label{eq:laplaces-approximation}
\end{equation}

The Kullback-Leibler divergence $\kullb{p}{q}$ between two distributions can be interpreted as the \textit{information gain} from $p$ over $q$; see \cite{Kullback1951}. The value of $\kullb{p(\bm c \condbar \dataset)}{p(\bm c)}$ will then represent the total amount of information about the context parameter distribution given by the dataset $\dataset$, compared to the prior, in the unit of nat. Multiplying this number with $\log_2(e)$, we get the information gain in bits.

Using Laplace's approximation, described in \eqref{eq:laplaces-approximation}, we can use the approximate posterior $q(\bm c \condbar \dataset)$ to calculate an estimate  $\kullb{q(\bm c \condbar \dataset)}{p(\bm c)}$ of the informativity of a dataset $\dataset$.

This procedure may be applied to the sequential learning setting considered later, where sequences of datasets are constructed by adding one new data point at a time. If the parameter space is well-behaved in the manner described above, one should expect the information gain to vary smoothly with dataset size.

In general, smoothness of the information gain does not guarantee that similar units have similar context parameters after calibration (the problem of estimating $c$ may be under-determined). However, it may indicate that the change in $p(\bm c \condbar \dataset)$ as a function of $\dataset$ is bounded. Thus, a smooth information gain suggests that units $k$ and $k+1$ which are similar in the sense that they give rise to similar datasets $\dataset_k$ and $\dataset_{k+1}$ \textit{can} at least be represented by context parameters $c_k$ and $c_{k+1}$ which are not too different.

\subsection{Context parameter dimension}
For a given unit, the context parameter $c$ can in principle take any value in $\mathbb R^K$. However, there is no guarantee that all degrees of freedom are used by the model. To better understand how the MTL architecture considered here learns to represent differences between units, one can analyze the \textit{intrinsic dimension} of $c$.

We follow \cite{Li2018}, and define the intrinsic context parameter dimension to be the smallest $k$ for which a context parameter confined to a random $k$-dimensional subspace of $\mathbb R^K$ achieves approximately the same performance as one allowed to take values in all of $\mathbb R^K$. Being particularly interested in the problem of few-shot learning by finetuning $\bm c$, we define the intrinsic dimension with respect to a model $(\hat{\bm c}, \hat{\bm \tau}, \hat \theta)$ which has already been trained.

To estimate intrinsic dimension in practice, we adapt the method described by \cite{Li2018} to the problem of finetuning $c$. The basis for this method is the definition
\begin{equation}
    c = c_0 + P c^{(k)}
    \label{eq:low-rank-context-variable}
\end{equation}
where $c_0 \in \mathbb R^K$ is a random, fixed vector, $P \in \mathbb R^{K \times k}$ a random, fixed, projection matrix, and the vector $c^{(k)} \in \mathbb R^k$ is learnable and initialized as zero. Dimension constrained learning can then be performed by freezing $\hat \theta$ and $\hat{\tau}$ and learning $c$ as defined by (\ref{eq:low-rank-context-variable}) with Algorithm \ref{alg:stochastic-gradient}. Since we have $p(c) = \normaldist(0, I)$, we simplify this procedure by defining $c_0 = 0$.

\section{Empirical study}
\label{sec:empirical-study}
As mentioned, it is commonly believed that data-driven VFMs require large amounts of data to perform well \cite{Bikmukhametov2020}. Thus, applying a data-driven VFM to wells with few observations provides an interesting case for few-shot learning. In this section, we apply our proposed few-shot learning strategy to the soft sensing problem of virtual flow metering on a large industrial dataset. The result is a VFM strategy which, given a sufficiently large dataset gathered from many wells, provides accurate predictions for new wells, even when the number of data points from the new well can be counted on one hand.

In the following, a single petroleum well constitutes a unit. The soft sensor that results from applying our proposed model as a VFM is similar to those used in \cite{Sandnes2021, Hotvedt2022} to perform multi-task learning across wells. Another relevant work is \cite{Grimstad2021}, which includes an investigation of noise models for neural network-based VFMs. However, the few-shot learning experiments conducted in this work consider a different learning problem and data availability regime than those presented in said works. Our model architecture also differs by its somewhat less problem-specific architecture as well as its inclusion of a learnable precision parameter which is unique for each unit. Finally, we conduct our study on a larger number of wells than previous studies. Studies in the literature typically concern the modeling of a single well \cite{Nemoto2023, Akhiiartdinov2023, Staff2020}. Our model architecture, in contrast, is applied jointly to 80 different petroleum wells. To our knowledge, our case study is the largest one in the VFM literature which uses real production data, in terms of the number of wells (including earlier work on multi-task learning \cite{Sandnes2021}).

\subsection{Learning problem}
For a single well at a given point in time, it may be possible to derive fairly simple relations between relevant sensor measurements and the total flow. The physics-based models which are used as the basis for VFMs in \cite{Nemoto2023, Akhiiartdinov2023, Staff2020, Hotvedt2022} fall into this category. For the instrumentation shown in Figure \ref{fig:well_sensors}, such relations often revolve around a model of choked flow which can be derived from Bernoulli's law:
\begin{equation}
    Q^{\textup{TOT}} = V(u) C \sqrt{\Delta p}
    \label{eq:bernoulli-model}
\end{equation}
Here, $Q^{\textup{TOT}}$ is the total volumetric well flow, $\Delta p$ is the pressure drop across the choke valve, $V(u)$ is a one dimensional function which depends on the choke valve geometry, and $C$ is a scalar which depends on properties of the fluid mixture which is produced from the well. The pressure drop $\Delta p$ is simple to calculate, and a model of the valve characteristic $V(u)$ may be available from the vendor of the choke valve. If $C$ were easy to estimate, this would then be a fairly simple problem of one dimensional regression. The reason why virtual flow metering is, by no means, a trivial problem, can largely be attributed to the complexity of modeling $C$. Due to the multi-phase nature of the flow which is being modeled, first principles-based modeling approaches typically consist of multiple nonlinear equations which must be jointly solved to provide an estimate of the fluid properties which in the case of (\ref{eq:bernoulli-model}) is represented by $C$ \cite{Sachdeva1986, Brill1999}. The model of $C$ will typically, similarly to the purely data-driven approach which will be used in the following case study, depend on observations of pressure, temperature, flow composition and gas lift rate. The optimal choice of equations for modeling $C$ is, however, strongly dependent on flow conditions, which varies between wells (and which may also vary over time for a given well). Furthermore, wells included in our case study may experience complex flow phenomena like slugging and critical flow, which may not be handled by steady-state first principles models. Thus, first principles-based approaches like those based on (\ref{eq:bernoulli-model}) are generally not applicable to the multi-unit dataset used here without significant manual work.

Our focus on data-driven models can be further motivated by \cite{Hotvedt2022}, which compares purely data-driven models based on multi-task learning with a mechanistic modeling approach of a similar (but more complex) structure to (\ref{eq:bernoulli-model}) in a multi-unit setting. Motivated by the superior performance of the MTL-based data-driven models reported by \cite{Hotvedt2022}, we do not investigate mechanistic or hybrid approaches in our case study. For further investigation of the advantages and disadvantages of a purely data-driven approach, comparisons with mechanistic models, and for an investigation of how these can be combined into a hybrid VFM, we refer to \cite{Hotvedt2022}.

\subsection{Dataset}
\label{sec:dataset}
\begin{figure}[bt]
\centering
\includegraphics[width=0.9\linewidth]{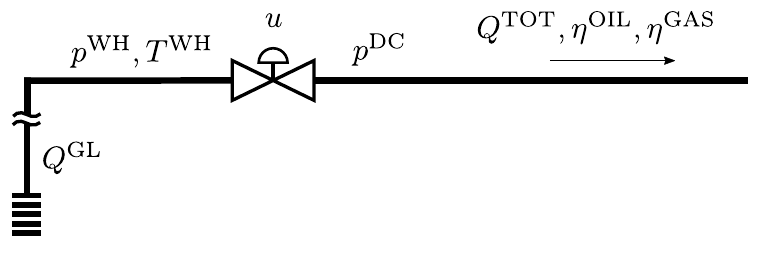}
\caption{Available measurements in a single petroleum production unit.}
\label{fig:well_sensors}
\end{figure}

\begin{figure}[bt]
    \centering
    \includegraphics[width=\linewidth]{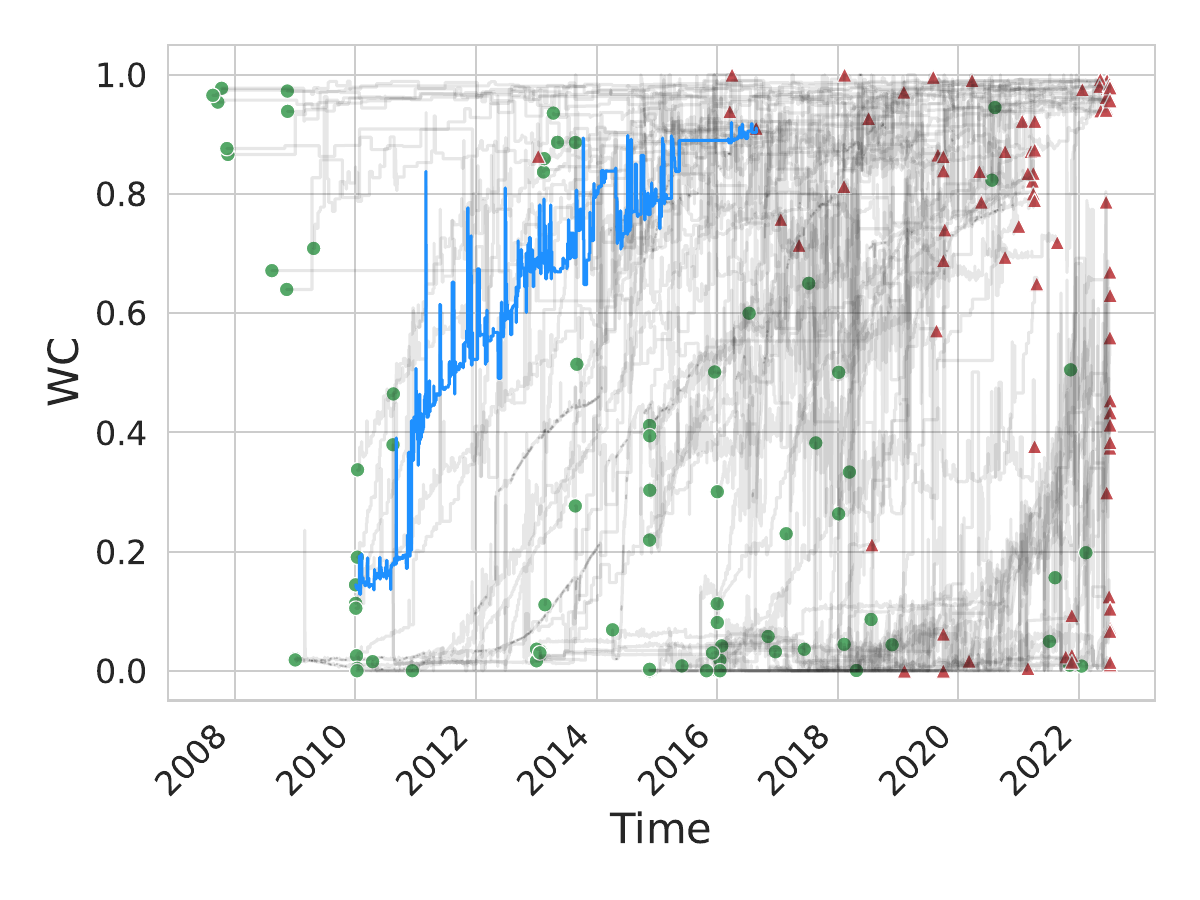}
    \caption{Water cut over time. The gray lines show the development for all 80 wells, and the blue line illustrates the general shape of these curves by showing the development for a single well.}
    \label{fig:fraction-timeseries}
\end{figure}

\begin{figure}[bt]
    \centering
    \includegraphics[width=\linewidth]{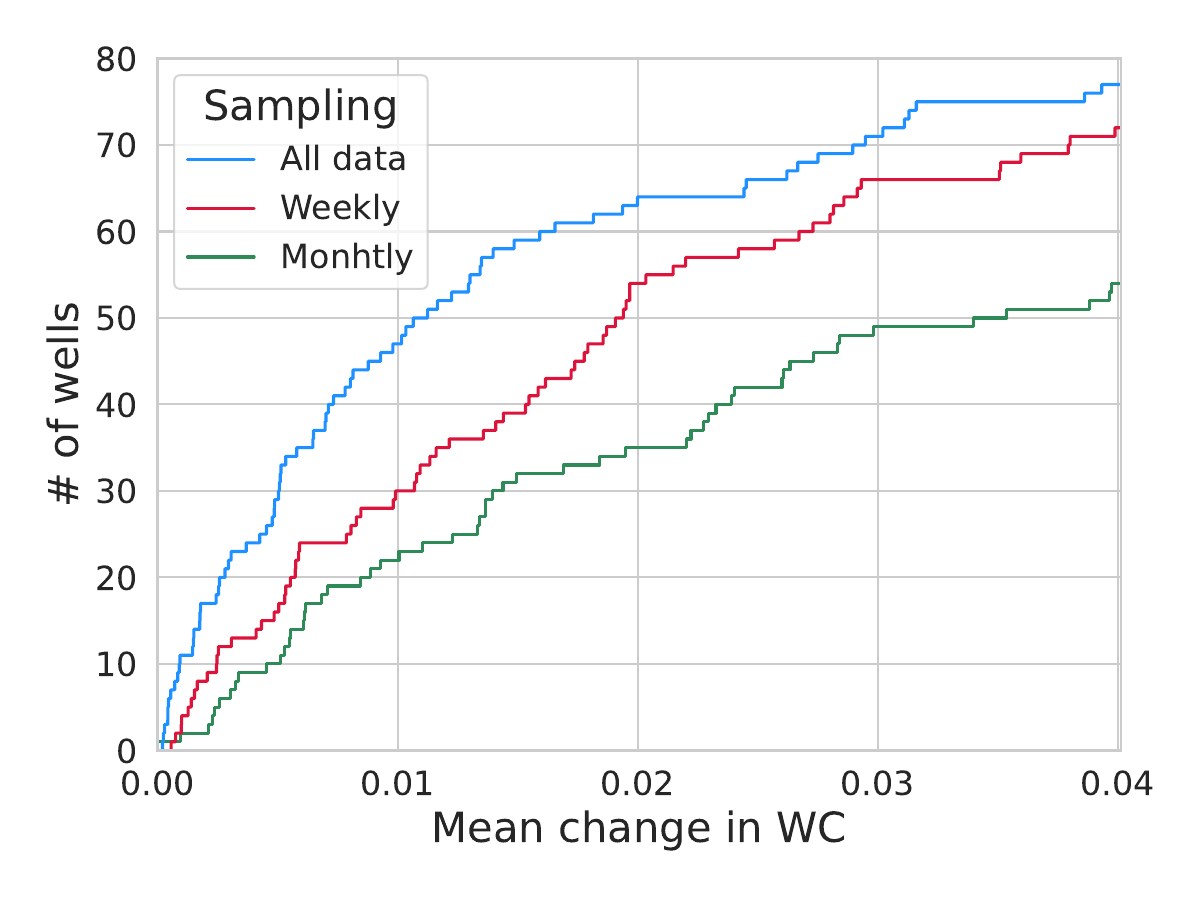}
    \caption{Cumulative distribution of change in water cut between data points, over all 80 wells.}
    \label{fig:fraction-change-distribution}
\end{figure}

\begin{figure}[bt]
    \centering
    \includegraphics[width=0.9\linewidth]{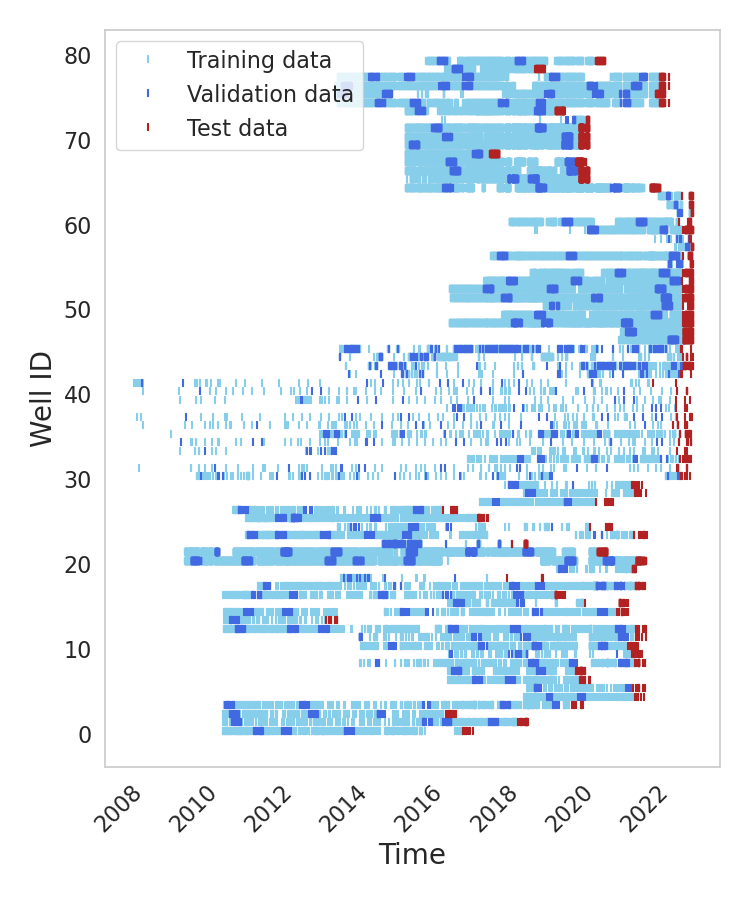}
    \caption{Partition of available data points into training (light blue), validation (dark blue) and test (red) sets. A rectangular marker indicates the availability of a single data point. Rectangles may overlap.}
    \label{fig:dataset}
\end{figure}

The dataset consists of 89 417 data points distributed across 80 petroleum wells from six oil fields, with instrumentation illustrated by Figure \ref{fig:well_sensors}. All wells are situated offshore, and produce a mixture of oil, gas and water. The data take the form of pairs $(x_{ij}, y_{ij})$ of observations, where
\begin{align*}
    x_{ij} & = [u_{ij},\, p^{\textup{WH}}_{ij},\, p^{\textup{DC}}_{ij},\, T^{\textup{WH}}_{ij},\, \eta^{\textup{OIL}}_{ij},\, \eta^{\textup{GAS}}_{ij},\, Q^{\textup{GL}}_{ij}], \\
    y_{ij} & = Q^{\textup{TOT}}_{ij}.
\end{align*}
Here, $u$ denotes choke valve opening, $p^{\textup{WH}}$ and $p^{\textup{DC}}$ denote pressures at the wellhead and downstream the choke valve, $T^{\textup{WH}}$ denotes wellhead temperature, $\eta^{\textup{OIL}}$ and $\eta^{\textup{GAS}}$ denote volumetric oil and gas fractions, $Q^{\textup{GL}}$ denotes volumetric gas lift rate and $Q^{\textup{TOT}}$ denotes total volumetric flow rate. Each data point consists of an average value of each measurement taken over an interval where flow is close to steady-state, detected using the technology described in \cite{Grimstad2016}. The observation frequency of these steady-state averaged data points varies depending on instrumentation, but generally appear at frequency on the order of days for MPFM measurements and weeks for well tests.

Out of the 80 wells, 69 employ gas lift and 41 have a multi-phase flow meter installed. Well tests are available for 61 of the wells. The basic system structure and available measurements are the same for all wells, while physical characteristics (like pipe lengths, diameters, friction coefficients and viscosities) differ between the wells. Thus, this dataset constitutes an example of multi-unit data, as described by Figure \ref{fig:multi-unit}.

Flow conditions also differ across the wells, and many wells are likely to experience complex flow phenomena like slugging or critical flow. Using the rule of thumb which states that critical flow happens when $P^{\textup{DC}} < 0.544 P^{\textup{WH}}$ \cite{Ros1960, Ashford1974}, we estimate that 67 of the 80 wells experience some critical flow, and 28 of the 80 wells experience critical flow for more than 25\% of the available observations. No filtering is done to remove data points subject to such effects, neither in the training nor in the test set.

Before processing, the observations were shifted and scaled to approximately lie in the unit interval. The measurements of total flow rate include both well tests and MPFM measurements, which our model could address through the use of separate precision parameters for each instrument type, as done in \cite{Grimstad2021}. However, for the sake of simplicity, we did not differentiate between the two types of data points in our experiments. We also refer to \cite{Grimstad2021} for a discussion of homoscedastic versus heteroscedastic noise models.

The oil and gas fractions $\eta^{\textup{OIL}}$ and $\eta^{\textup{GAS}}$ are only available from well tests and MPFM measurements. Thus, online observations of these variables will generally not be available to the model when it is used as a VFM. This may cause problems, because fractions tend to change significantly over time. The drift in fractions is apparent from Figure \ref{fig:fraction-timeseries}, which shows how the \textit{water cut} (WC), i.e. the ratio between flow rate of water and the total flow rate of all liquids, develops over time for the dataset used in this case study. As indicated by the figure, the water cut typically starts low (often close to zero) and increases steadily during the lifetime of the well, ending up at a high level (often approaching $100\%$). The gas fraction $\eta^{\textup{GAS}}$ also tends to vary over time in a similar, but somewhat less monotonic manner.

However, it is worth noting that the change in fractions over time tends to be slow. This is illustrated by Figure \ref{fig:fraction-change-distribution}, which shows the cumulative distribution (over all wells in the dataset) of the absolute change in water cut between subsequent data points. The blue line shows this distribution for all data, and the red and green lines show the distributions for data which have been down-sampled to have observation frequencies of minimum a week and a month, respectively. For the complete dataset, the approximation error introduced by using the previous available well test seems to be relatively small (for over 60 of the 80 wells, the mean change in WC is below 0.02). The error increases as the observation frequency goes down, but is still fairly low even in the case of monthly observations (where over half of the wells have mean changes below 0.025). In an online implementation of the methods proposed here, one should expect the error introduced by a zeroth-order approximation where one sets the model inputs $\eta^{\textup{OIL}}$ and $\eta^{\textup{GAS}}$ equal to their value at the last available measurement to be relatively small, as long as the observation frequency is not too low. When the observation frequency decreases, however, one should expect model performance to deteriorate accordingly.

Figure \ref{fig:dataset} shows a split into training, validation and test set which was done using the method described in \cite{Sandnes2021}. The method, which splits the dataset into smaller chunks of time to reduce overlap between the sets, results in 81\%, 14\% and 5\% of the data points being assigned to the training, validation and test sets, respectively. As can be seen from the figure, the wells present in the dataset cover a range of different data frequencies and time spans.

\begin{table}[bt]
    \centering
    \begin{tabular}{lrrrrr}
    \hline
                       &   P10 &   P25 &   P50 &   P75 &   P90 \\
    \hline
     CHK median        &  0.35 &  0.44 &  0.58 &  0.99 &  1.00 \\
     CHK IQR           &  0.00 &  0.07 &  0.15 &  0.29 &  0.47 \\
     CHK range         &  0.39 &  0.51 &  0.73 &  0.81 &  0.86 \\
     Pres. drop median &  0.01 &  0.02 &  0.10 &  0.25 &  0.35 \\
     Pres. drop IQR    &  0.01 &  0.01 &  0.09 &  0.17 &  0.41 \\
     Pres. drop range  &  0.11 &  0.27 &  0.44 &  0.60 &  0.71 \\
     GLR median        &  0.01 &  0.04 &  0.11 &  0.20 &  3.18 \\
     GLR IQR           &  0.00 &  0.01 &  0.04 &  0.11 &  0.91 \\
     GLR range         &  0.04 &  0.10 &  0.19 &  0.40 &  4.01 \\
    \hline
    \end{tabular}
    \caption{Summarizing statistics for case study dataset. CHK denotes choke valve opening and GLR denotes gas-to-liquid ratio, both of which are fractions which take values in the unit interval. Pressures are initially given in bar, but are scaled down by a factor of 100 to approximately lie in the unit interval.}
    \label{tab:dataset-summary}
\end{table}

Table \ref{tab:dataset-summary} provides a high-level summary of the dataset by reporting the distribution of and variation in choke valve opening, pressure drop and \textit{gas-to-liquid ratio} (GLR). The table puts particular emphasis on inter-well variation, and reports the 10th, 25th, 50th, 75th and 90th percentiles \textit{across wells} of the per-well median, inter-quartile range (IQR) and total range. As can be seen from this table, both operating conditions and variation in operating conditions differ significantly between wells. For instance, the per-well IQR of the choke valve has a lower quartile value (across wells) of 7\%, while its upper quartile is situated at 29\%. The lower and upper quartiles of the per-well pressure drop range lie at 27 bar and 60 bar, respectively. The per-well median of the gas-to-liquid ratio has lower and upper quartiles at 0,04 and 0,20.

For an in-depth analysis of additional data challenges (including problems relating to well depletion, operating practices and resulting covariate correlation) for a dataset which is largely overlapping with ours, we refer to Chapter 4 of \cite{Sandnes2024}.

\subsection{Pretraining}
\label{sec:convergence-rate-experiment}
The model parameters $(\bm c, \bm \tau, \theta)$ can be estimated from data by using the multi-task learning gradient ascent method in (\ref{eq:sgd}), or another stochastic gradient-based method, with gradients estimated by Algorithm \ref{alg:stochastic-gradient}. Since the dimension of $\theta$ exceeds the dimensions of $\bm c$ and $\bm \tau$ by orders of magnitude for our model, one can expect that most of the complexity of this learning problem can be attributed to the estimation of gradients with respect to $\theta$. Thus, it would be convenient to identify settings in which one can justify freezing $\theta$, instead only estimating $(\bm c, \bm \tau)$.

To guide the search for such conditions, one can consider the theoretical convergence rate of the MAP estimator in the context of MTL. Assuming that the units (which in this case define tasks) are sufficiently similar, and that each unit contributes with a comparable number of data points that does not change with $M$, the theoretical expected learning rate as a function of the number $M$ of units in the dataset will, in a large number of multi-task learning settings, be $\mathcal{O}(1 / \sqrt{M})$ \cite{Zhang2021}. Since this function flattens out as $M$ increases, one should expect diminishing returns from each single new well as their number increases.

We tested this assumption by training and estimating the performance of an MTL model on data from $M = 1, 5, 10, 20, 40, 80$ units. For each value of $M$, the dataset was randomly divided into $80/M$ disjoint groups of $M$ different units, resulting in an MTL dataset with $M$ distinct tasks. Then, a model was learned from the training data shown in Figure \ref{fig:dataset} corresponding to each group. The neural network was chosen to have large capacity considering the problem complexity; see the appendix for additional implementation details.

After training, the model was evaluated on a previously unseen test set drawn from the same wells as its training set (as illustrated by Figure \ref{fig:dataset}). This procedure was repeated 20 times with different random seeds to reduce noise due to well group partitioning, and from stochasticity related to neural network training.

\begin{figure}[bt]
    \centering
    \includegraphics[width=\linewidth]{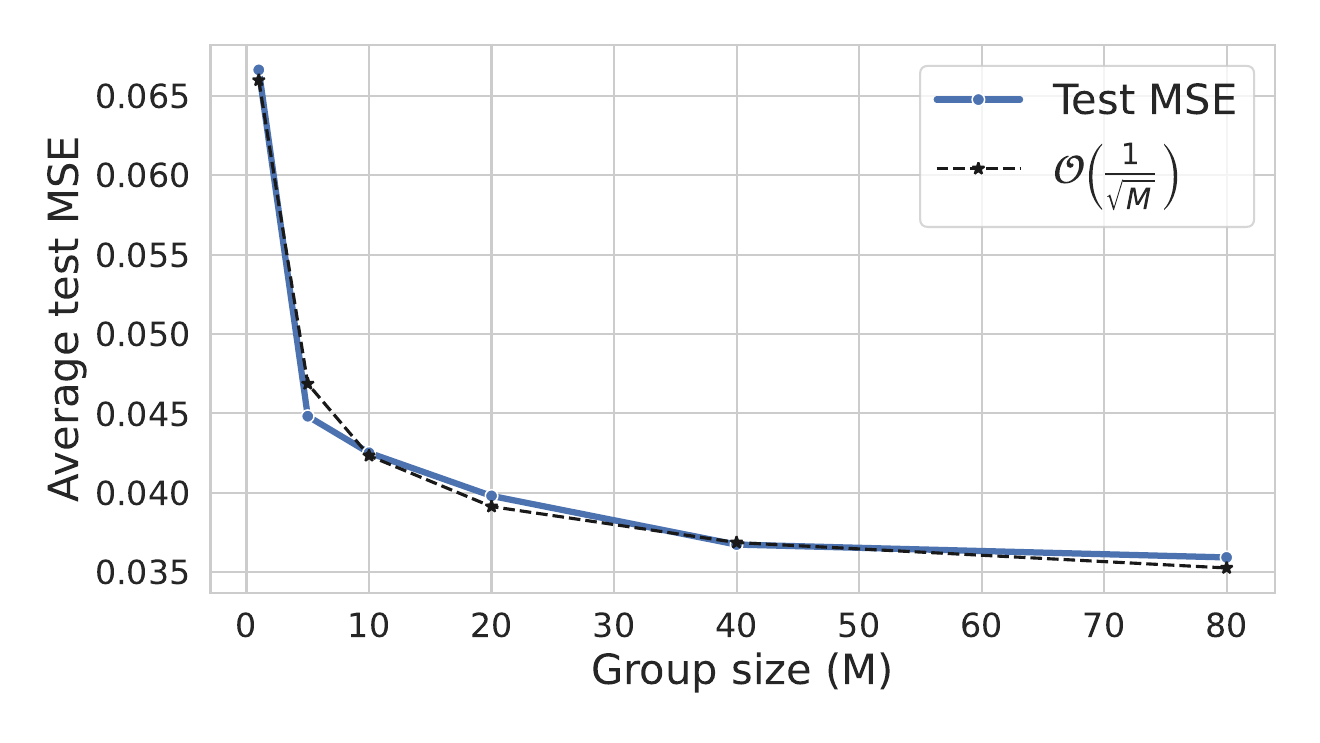}
    \caption{MTL performance for increasing number of wells.}
    \label{fig:mtl-scaling-example}
\end{figure}

Figure \ref{fig:mtl-scaling-example} shows the test set MSE, averaged over all wells and experiment repetitions, as a function of the number of units from which data were available during training. In addition, the figure shows a function in $\mathcal{O}(1 / \sqrt{M})$ fitted to this curve using least squares regression. From the figure, one can see that the theoretically predicted convergence rate lies close to the observed one. Furthermore, one can see that the prediction error flattens out significantly between 40 and 80 wells.

Averages over wells for each individual experiment run are shown in Figure \ref{fig:mtl-scaling-example-all-runs} in the appendix.

\subsection{Few-shot learning performance}
\label{sec:few-shot-experiment}
\begin{figure}[bt]
    \centering
    \includegraphics[width=\linewidth]{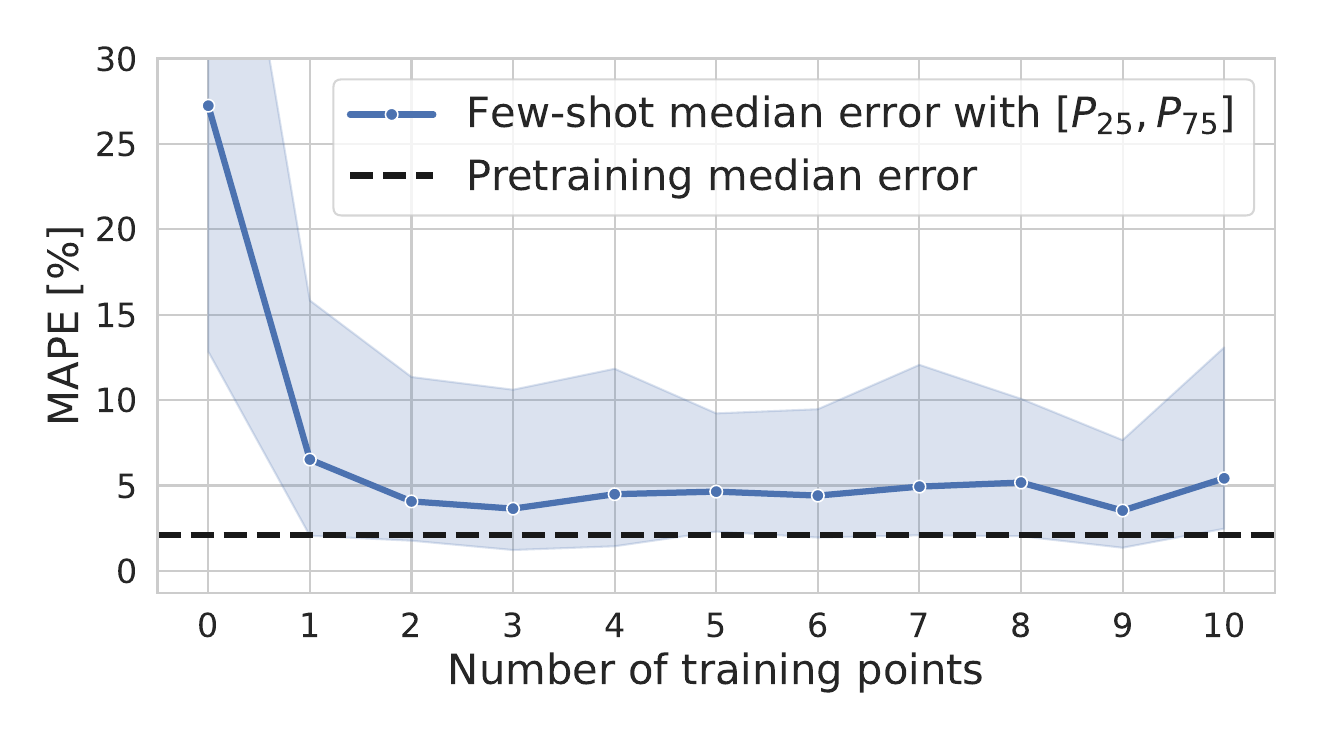}
    \caption{Few-shot learning performance. The x-axis shows the number of available data points for the wells on which few-shot transfer learning is performed. The pretrained models had access to an average of 903 training data points from each well.}
    \label{fig:few-shot-results}
\end{figure}

In this case study, multi-task pretraining is mainly a means to achieve few-shot transfer learning; our main objective is to demonstrate the use of few-shot learning to achieve high data-driven VFM performance in data-scarce settings, by performing fine-tuning from an MTL model learned from a larger dataset. To this end, the wells were partitioned into two sets: The set of \textit{base wells}, which provide many data points and are used for MTL model pretraining, and the set of \textit{holdout wells}, which provide few data points and are used for few-shot transfer learning by context parameter fine-tuning.

For the holdout wells, the data could not be used as is, since most of the wells shown in Figure \ref{fig:dataset} are arguably not suffering from data scarcity (in particular, this is true for the wells equipped with an MPFM). To simulate a low-data setting for the holdout wells, we used a method where
% The intention of the few-shot learning experiment was to simulate a data-scarce setting where new measurements appear at regular but infrequent intervals. Since this is not the case for the dataset used in the preceding section, the few-shot learning experiment required data to be drawn using a different strategy from the one illustrated by Figure \ref{fig:dataset}. The strategy which was used simulated a data-scarce setting in which
measurements appear one at a time, sequentially in time, with at least one week in between. For a more detailed description of the dataset generation method, see the appendix.

In the first step of the experiment, the multi-task learning procedure described in Algorithm \ref{alg:stochastic-gradient} was used to learn a base model from a training set consisting of all available data from 60 randomly chosen base wells. The reason for choosing this particular number of base wells was largely motivated by the need for a holdout set of a size sufficient for robust evaluation of model performance. However, we argue that $M=60$ is situated safely in what can be considered a flat part of Figure \ref{fig:mtl-scaling-example}, justifying freezing the value $\theta$ learned from the 60 base wells.

Pretraining resulted in a base model with parameters $(\hat{\theta}, \hat{\bm c}, \hat{\bm \tau})$. This model was used in the succeeding transfer learning step, where calibration was performed for each of the 20 holdout wells $k > M$. Here, the neural network parameters $\theta = \hat{\theta}$ were fixed at their value from pretraining, the precision parameter was kept fixed at $\hat{\tau}_k = (\hat{\tau}_1 + \cdots + \hat{\tau}_M) / M$, while the context parameter $c_k$ was calibrated to datasets of increasing size from the holdout well (from 0 up to 10 data points), using our proposed finetuning strategy. Further implementation details are given in the appendix.

\begin{figure}[bt]
    \centering
    \includegraphics[width=\linewidth]{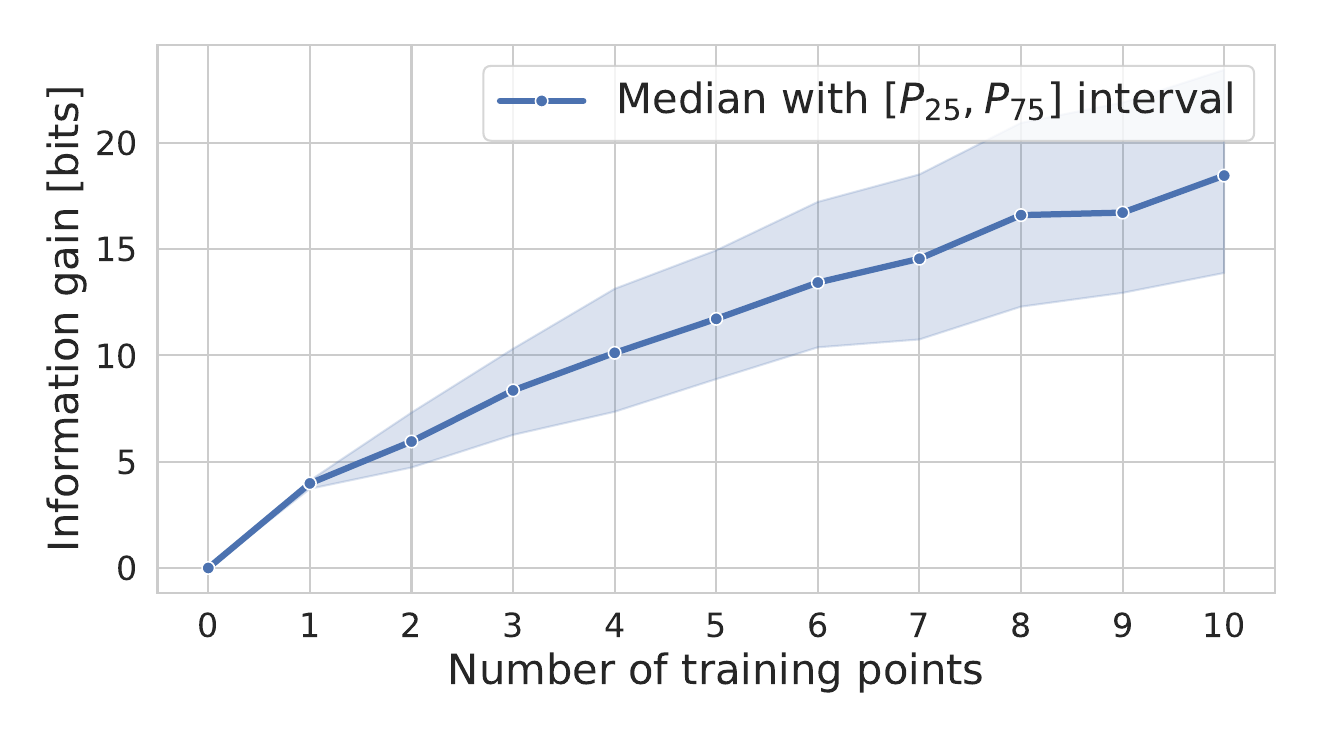}
    \caption{Information gain for training datasets of increasing size. The x-axis shows the number of available data points for the wells on which few-shot transfer learning is performed.}
    \label{fig:information-gain}
\end{figure}

To increase the statistical strength of the experiment, this procedure was repeated 20 times with different random choices of the 60 base wells. This resulted in each of the 80 wells appearing in the set of holdout wells at least once. To enable cross-unit comparison, a randomly chosen experiment run was chosen to represent the few-shot learning performance of the calibrated model on each well.

Figure \ref{fig:few-shot-results} describes the development of the mean absolute percentage error (MAPE) on the test set as the number of available data points increases. Here, the blue line shows the median test MAPE across wells, and the colored region shows 50\% probability intervals over all wells. The black dashed line indicates the 2.11\% MAPE achieved by the base model when evaluated in the same way as the few-shot learning model on the first week of the test set shown in Figure \ref{fig:dataset}.

\subsection{Analysis of context posteriors}
Figure \ref{fig:information-gain} describes the information gain $\kullb{q(\bm c \condbar \dataset_k)}{p(\bm c)}$ for holdout datasets $\dataset_k$ of increasing size, as estimated by the Laplace approximation procedure described above. The blue line shows the information gain as a function of dataset size, while the shaded blue region shows 50\% probability intervals calculated across wells. One can see from the figure that dataset information gain increases steadily with dataset size. Furthermore, the slope of the curve seems to be decreasing, indicating that the effect of additional data points is diminishing.

\subsection{Context parameter dimension analysis}
\begin{figure}[bt]
    \centering
    \includegraphics[width=\linewidth]{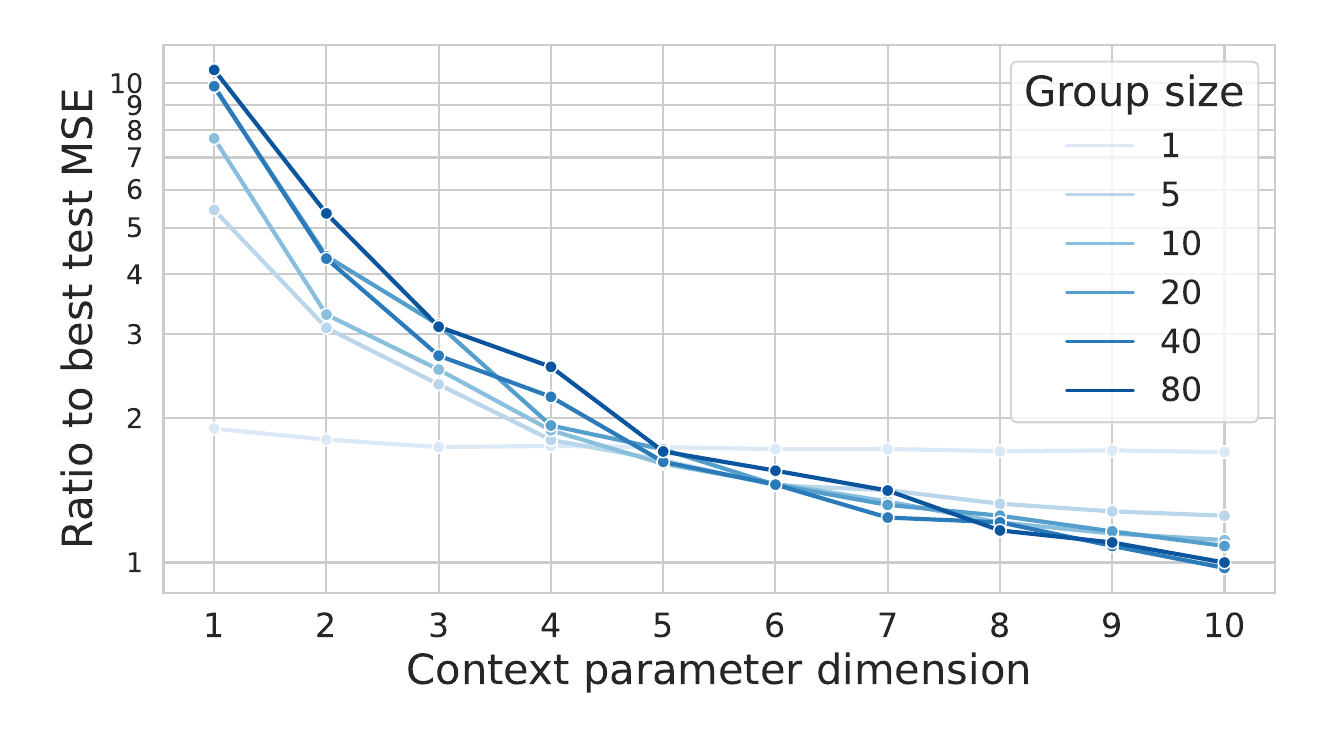}
    \caption{Scaled test set MSE for different context parameters dimensions. Values are scaled by the MSE achieved for dimension 10 (0.0378), which is comparable to the MSE achieved for $M=80$ in the pretraining experiment (0.0359). Note the logarithmic scaling of the y-axis.}
    \label{fig:intrinsic-dimension}
\end{figure}

The role of the context parameter dimensionality was analyzed by exchanging the context parameter of the pretrained MTL models with context parameters described by (\ref{eq:low-rank-context-variable}), and performing intrinsic dimension estimation as described above. This was done for all 20 pretraining experiment repetitions, with a new projection matrix $P$ being initialized for each well and experiment repetition.
% , and results were averaged over repetitions. For each run, a different projection matrix $P$ was initialized.

Figure \ref{fig:intrinsic-dimension} shows the resulting test set MSE, averaged over wells and repetitions, scaled as a fraction of the test set MSE achieved with dimension $k = K = 10$, for different group sizes. For all $M > 1$, the plot shows that the first few dimensions make the largest contributions to reducing test set error, which flattens out somewhat around dimension 5-6. Still, it is worth noting that the y-axis starts at a fraction which represents rather poor model performance (10 times the best MSE), and that test set error significantly decreases from dimension 6 and all the way up to the full 10 dimensions. Finally, the number of dimensions needed to capture the variation between wells grows with the number of wells which is modeled simultaneously. In the single-unit case $M = 1$, the context parameter seems to be of little use.

\section{Concluding remarks}
\label{sec:conclusion}
Figure \ref{fig:mtl-scaling-example} suggests that the convergence rate of $\mathcal{O}(1 / \sqrt{M})$, where $M$ is the number of units, applies well to the VFM problem. The flatness of this convergence rate for large values of $M$ serves as motivation for our proposed few-shot learning approach, where a pretrained MTL model with fixed parameters acts as a base model. The result is a data-efficient learning algorithm, where only the low-dimensional context parameters are calibrated to data from a new unit.

From Figure \ref{fig:information-gain}, one can see that the dataset information gain varies smoothly with dataset size. This indicates that the regularization which is introduced by the prior distribution over the context variable reduces the complexity of the learned context parameter space. The smoothness of the parameter space as well as its low dimension suggests that the learned context parameters may be meaningful to the model, even if they do not permit a physical interpretation. Furthermore, the results shown in Figure \ref{fig:intrinsic-dimension} indicate that even though a few dimensions accounts for a large part of the modeled inter-unit variation, all $K = 10$ context parameter dimensions are exploited to provide a more refined representation of the variation between wells; arguably then, the effective dimension of the context parameter (in the context of finetuning) is $k=K=10$. Given the regularizing role played by the context parameter, this is perhaps not too surprising.

Our main result is shown in Figure \ref{fig:few-shot-results}. From this figure one can see that our proposed calibration procedure only requires a handful of training data points for the resulting model to achieve high performance on a previously unseen well -- the median error across new units drops below 5\% after calibration to only three data points, which can be considered as a high performance for a VFM \cite{Amin2015,Grimstad2021}. There is still a gap to the performance of about 2,5\% error achieved in the pretraining setting, as indicated by the dashed line in Figure \ref{fig:few-shot-results}. We hypothesize that this is due to the \textit{non-stationarity} of the petroleum production process \cite{Hotvedt2022}; over time, a petroleum well depletes and production equipment is subject to wear, meaning the number of up-to-date data points is limited by data observation frequency. Since this frequency is lower in the few-shot setting considered here (where a new data point is observed every week) than in the multi-unit pretraining setting (where about half of the wells have MPFMs and may get access to new data points multiple times a week), a performance gap is to be expected.

It should be noted that the high performance of the fine-tuned VFM relies on the existence of a good pretrained MTL model. One could argue that we are then not in a data-scarce setting, since successful pretraining requires a base dataset with data from a quite large number of well-instrumented wells. However, we emphasize that this pretraining only has to happen \textit{once}. As long as one has access to a fixed set of wells that give rise to high-quality training data, the few-shot transfer learning methodology can be used to transfer this information to any other well of the same type. %In a sense, the worse instrumented wells can ``borrow" the instrumentation of the high data-quality wells.

It should also be noted that since the learning problem is highly heterogeneous, the reported prediction errors are aggregated over a large range of different operating conditions and flow regimes. We emphasize that the reported results are an average, and that model performance may co-vary with all of the conditions which results are aggregated over. Even though our results indicate strong model performance on average, it would be interesting to investigate in further detail under which conditions the proposed methodology is the most useful, and under which conditions it struggles. It would be particularly interesting to investigate how model performance is correlated with flow regime. Such an investigation is, however, not straight-forward to conduct, since flow regime classification constitutes its own soft sensing problem. Addressing this soft sensing problem would be an interesting topic for future research, and a solution to this problem could be used both to improve assessment of VFM performance, and to improve the performance of the VFM itself (by providing additional contextual information, e.g. in the form of an additional, data-dependent context variable).

As illustrated by the few-shot learning experiment, adaptation of the pretrained model to previously unseen wells requires little or no extra manual work. In contrast, the effort required to apply a first principles VFM in the same setting (in which the model must be adapted to 20 new wells) would be prohibitively large. Thus, our proposed fine-tuning strategy may enable the use of VFMs in low-data settings that were previously deemed impractical, e.g. due to insufficient budgets for VFM deployment and maintenance.

In conclusion, we have demonstrated empirically that our multi-unit VFM can model a large range of wells, and that it can provide accurate predictions for new wells after calibration to only a handful of observations -- not unlike a first principles VFM.

\subsection{Acknowledgments}
This work was supported by Solution Seeker AS.

\subsection{Conflict of interest}
Authors state no conflict of interest.

\printbibliography
\appendix

\section{Appendix}
\label{app:implementation-details}
This section describes in further detail how the model and learning algorithm were implemented and evaluated in the case study.

\subsection{Model}
\begin{figure}[bt]
    \centering
    \includegraphics[width=\linewidth]{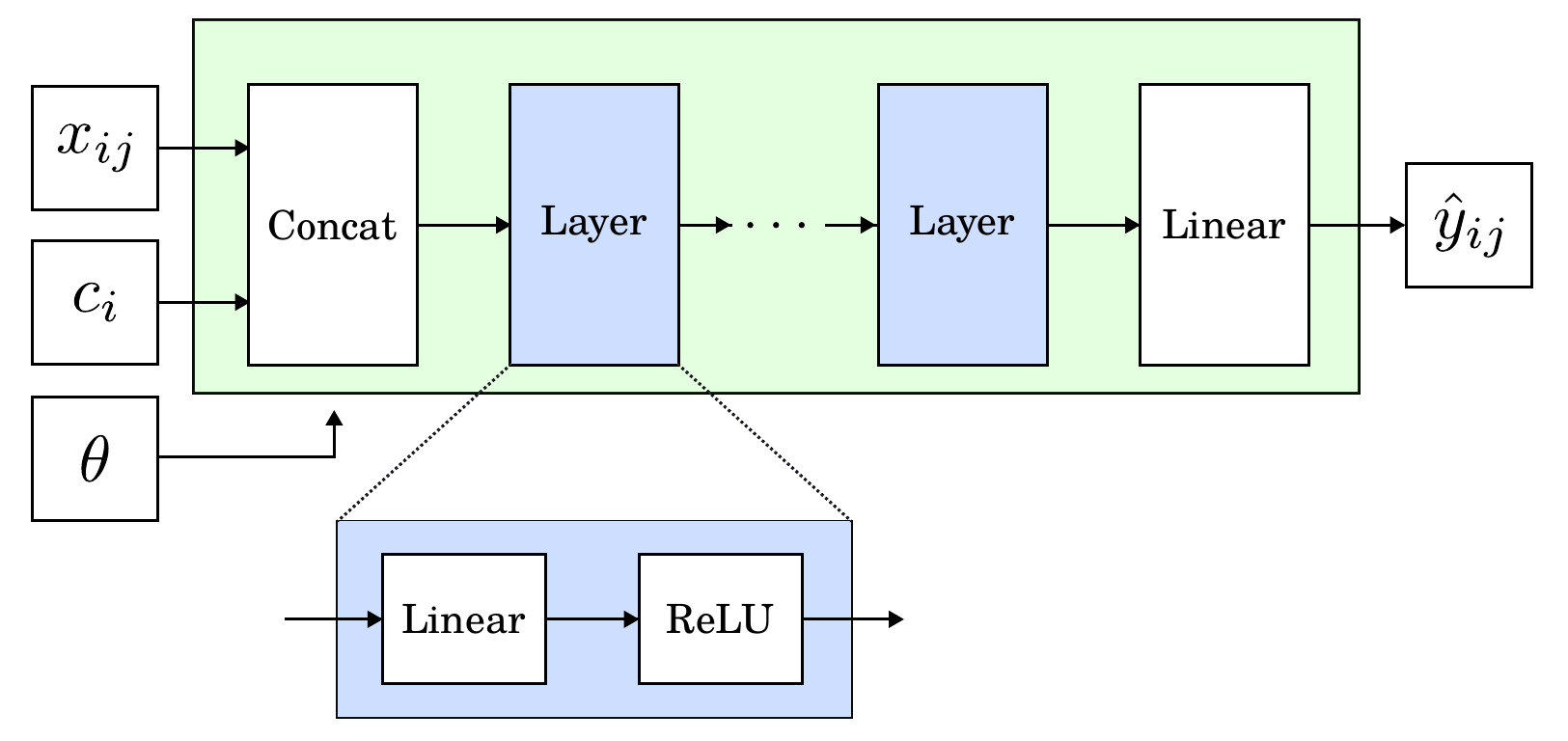}
    \caption{Neural network architecture implementing $f(x; c, \theta)$. A data point $x_{ij}$ from unit $i$ is processed to produce a prediction $\hat{y}_{ij}$. $\theta$ is the parameters of the linear layers in the network and $c_i$ is the context parameters of unit $i$ (whose data is being processed). The precision parameter $\tau_i$ is not shown.}
    \label{fig:nn-architecture}
\end{figure}

The multi-unit soft sensor was modeled using a deep neural network, implemented in PyTorch \cite{Pytorch2019}. The network architecture is illustrated in Figure \ref{fig:nn-architecture}. It consists of a feed-forward neural network of width 400 and depth 4, which has been shown to be sufficiently large in similar studies on neural network-based VFMs (see e.g. \cite{Sandnes2021}, \cite{Hotvedt2022}). It uses ReLU activation functions, and its input is the concatenation of the input data point $x_{ij}$ and a learnable, unit-specific context vector $c_i$ of dimension 10. The context vectors of the units can be thought of as learned embedding vectors. Given $x_{ij}$, the neural network returns an estimate $\hat{y}_{ij} = f(x_{ij}; c_i, \theta)$ of the mean of $y_{ij}$. Together with a precision parameter acquired by passing the learnable unit-specific parameter $t_i$ through the softmax function $g$, this estimate is used to form the conditional distribution
\begin{equation*}
    p(y_{ij} \condbar x_{ij}, c_i, t_i, \theta_i) = \normaldist(y_{ij} \condbar f(x_{ij}; c_i, \theta), 1 / g(t_i)).
\end{equation*}

\subsection{Pretraining}
During pretraining, all model parameters $(\bm c, \bm t, \theta)$ were learned jointly through the mini-batch gradient ascent procedure described in Algorithm \ref{alg:stochastic-gradient}, using the Adam optimizer \cite{Kingma2015} with a batch size of 2048. For both the convergence rate experiment described and the few-shot learning experiment, training was run for 20 000 epochs with a learning rate of $1 \cdot 10^{-4}$. In both cases, the model with the lowest validation set MSE was returned. The training was conducted on a desktop computer with a single GPU.

\subsection{Calibration}
Context parameters $c_k$ were calibrated using data from previously unseen units $k > M$. For each new data point, the context parameter $c_k$ was reinitialized and training was done from scratch. This training consisted of 100 epochs of stochastic gradient ascent with a learning rate of $1 \cdot 10^{-4}$. Due to the low data availability, no model selection procedures were performed in this setting.

\subsection{Model validation in few-shot learning}
\label{app:few-shot-data-split-algorithm}
The dataset shown in Figure \ref{fig:dataset} consists of data from wells with significant differences in instrumentation and practice of operation. As can be seen in Figure \ref{fig:dataset}, this results in significant differences between wells in data frequency as well as dataset duration. To increase the comparability of results across wells, we devised a procedure inspired by how new well tests typically arrive in a petroleum production setting. The result was the following procedure, which was used for each of the unseen wells $k > M$:
\begin{enumerate}
    \item Set $\dataset_{k} = \emptyset$
    \item For $j=1, \dots, 10:$
    \begin{enumerate}
        \item Set $(x_{kj}, y_{kj})$ to be the next available data point which is situated a week or more into the future, and add $(x_{kj}, y_{kj})$ to the dataset $\dataset_{k}$
        \item Calibrate $c_k$ to the dataset $\dataset_{k}$
        \item Evaluate the prediction error of the model on all available data from the week following the latest data point in $\dataset_{k}$. If no data is available, use the single next available data point as test set. 
    \end{enumerate}
\end{enumerate}
In this procedure, the division into training, validation and test sets shown in Figure \ref{fig:dataset} was ignored, and all data from each given well was used when running the procedure described above.

The relation between available measurements and the outflow of an oil well is known to change significantly over time \cite{Hotvedt2022}, which is why the test set is limited to have a duration of one week. This results in the test set size differing between wells. However, the error metrics used take the form of averages, which means that test set size does not affect the magnitude of the error metric for a given well. Thus, we argue that the aggregation of results done later can be justified, even though the statistical strength of the test set error may differ between wells due to differing test set sizes.

\subsection{Additional results}
\label{app:additional-results}
\begin{figure}[bt]
    \centering
    \includegraphics[width=\linewidth]{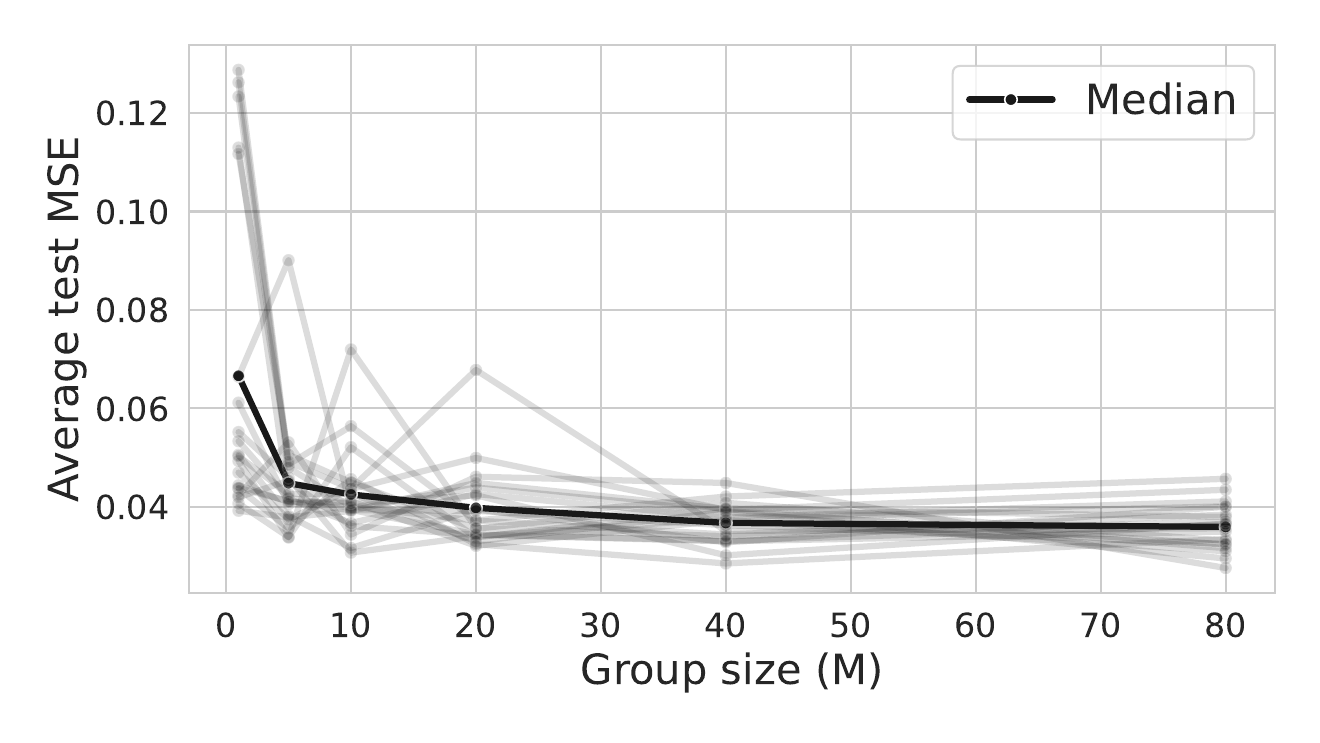}
    \caption{MTL performance for increasing number of $M$. The x-axis shows the number of available data points for the wells on which few-shot transfer learning is performed. Median across runs is shown in black, individual runs are shown in gray.}
    \label{fig:mtl-scaling-example-all-runs}
\end{figure}

Figure \ref{fig:mtl-scaling-example-all-runs} shows each of the 20 individual pretraining experiments. The variance observed between individual experiments supports the large number of experiment repetitions performed when generating Figure \ref{fig:mtl-scaling-example}.

\end{document}